\PassOptionsToPackage{table}{xcolor}
\documentclass[11pt]{article}

\usepackage[final]{acl}

\usepackage{times}
\usepackage{latexsym}

\usepackage[T1]{fontenc}

\usepackage[utf8]{inputenc}

\usepackage{microtype}

\usepackage{inconsolata}

\usepackage{graphicx}

\usepackage{graphicx}
\usepackage{tabularx}
\usepackage{booktabs} 
\usepackage{wrapfig}
\usepackage{amsmath}
\usepackage{amsfonts}
\usepackage{colortbl}   
\usepackage{CJKutf8}
\usepackage{booktabs}
\usepackage{multirow}
\usepackage{geometry}
\usepackage{pifont} 
\usepackage{makecell} 
\usepackage{array}
\usepackage{float}
\usepackage{tcolorbox}
\usepackage{stfloats}

\newcommand{\cmark}{\ding{51}}%
\newcommand{\xmark}{\ding{55}}%
\newcommand{\win}[1]{\colorbox{green!25}{{#1}}}
\newcommand{\loss}[1]{\colorbox{red!20}{{#1}}}

\title{Enhancing Multilingual Reasoning via Steerable Model Merging}

 \author{Zhuoran Li$^{1}$, Rui Xu$^{2}$, Jian Yang$^{3}$, Junnan Liu$^{4}$, Zhijun Chen$^{3}$, Qianren Mao$^{5}$\\
  {\bf Hongcheng Guo$^{2}$,  Jiaheng Liu$^{6}$, Likang Xiao$^{3}$, Ming Li$^{7}$, Xiaojie Wang$^{1}$\thanks{$\ \ $Corresponding author}} \\
    $^1$Beijing University of Posts and Telecommunications, $^2$Fudan University\\
    $^3$ Beihang University, $^4$ Monash University, $^5$Zhongguancun Laboratory\\
    $^6$Nanjing University, $^7$Tsinghua University
     }

\begin{document}
\maketitle
\begin{abstract}
Model merging is an effective technique for composing the capabilities of a multilingual model and a reasoning model. It has achieved promising generalization in multilingual reasoning tasks by aligning feature spaces of different models. However, the merged single model often fails to address the conflicts between source models, leading to suboptimal performance. In other words, the one-size-fits-all merging strategy may not align with the characteristics of different inputs which may require prioritizing certain models over others. To this end, we propose a Steerable Model Merging (\textbf{ST-Merge}) framework to modulate the contribution of each source model. To realize this idea, we introduce a gated cross-attention mechanism to weight or filter the two attended source models in an adaptive manner. Extensive experiments demonstrate that ST-Merge consistently outperforms multiple strong baselines on four multilingual reasoning benchmarks across 21 different languages.

\end{abstract}

\section{Introduction}

\begin{figure}
  \centering
  \includegraphics[width=0.41\textwidth]{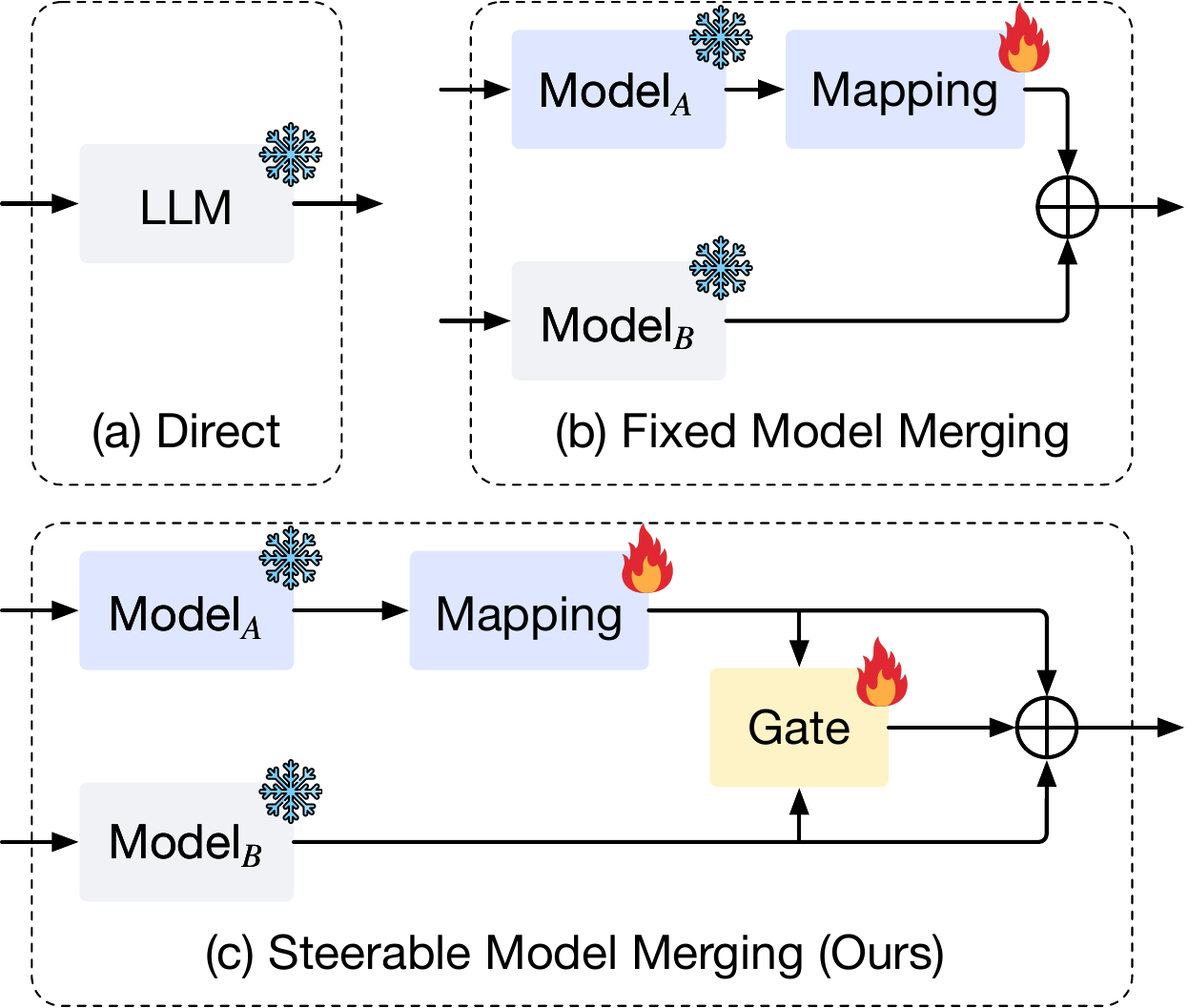}
  \caption{Illustration of our ST-Merge idea. (a) Direct application of LLM to all languages. (b) One-size-fits-all model merging method. (c) The proposed steerable model merging method learns to modulate the contribution of each source model for different inputs. LLM/$\text{Model}_B$: the reasoning LLM. $\text{Model}_A$: the external multilingual encoder.}
  \label{fig:example}
\end{figure}

Multilingual reasoning aims to empower Large Language Models (LLMs) to perform complex reasoning tasks across diverse languages. This capability is valuable in circumstances where limited or no annotations are available for low-resource languages. In recent years, reasoning large language models, such as MetaMath~\cite{DBLP:conf/iclr/YuJSYLZKLWL24/metamath} and Orca~\cite{DBLP:journals/corr/abs-2311-11045/orca2}, have achieved significant performance improvements through parameter-efficient fine-tuning on source language data and direct application to target language data (as shown in Figure~\ref{fig:example}(a)). 

Furthermore, it has been discovered that additional multilingual representation alignment improves the low-resource language reasoning performance by composing an external multilingual encoder to replace or augment the original LLM query embedding. This strategy, named Model Merging, has demonstrated improvements in many multilingual reasoning tasks~\cite{yoon-etal-2024-langbridge, NEURIPS2024_3bf80b34/mindmerger} (as shown in Figure~\ref{fig:example}(b)).

However, as a one-size-fits-all strategy, current fixed model merging approaches struggle to strike an optimal balance for inputs across diverse languages. On one hand, over-reliance on the external multilingual encoder can dilute the core reasoning capabilities inherent to the original LLM, potentially leading to \textit{catastrophic forgetting}. On the other hand, insufficient reliance on the external encoder hampers the understanding of low-resource languages, thereby limiting reasoning performance. Existing studies have observed that fixed model merging often causes a degradation in reasoning capabilities for languages in which the LLM is already proficient~\cite{yoon-etal-2024-langbridge}. Therefore, it is imperative to devise an adaptive scheme to modulate the models based on the characteristics of inputs.

\begin{figure}[!t]
\centering
\begin{tabular}{ccc}
    \hspace{-0.35cm} \includegraphics[width=0.32\linewidth]{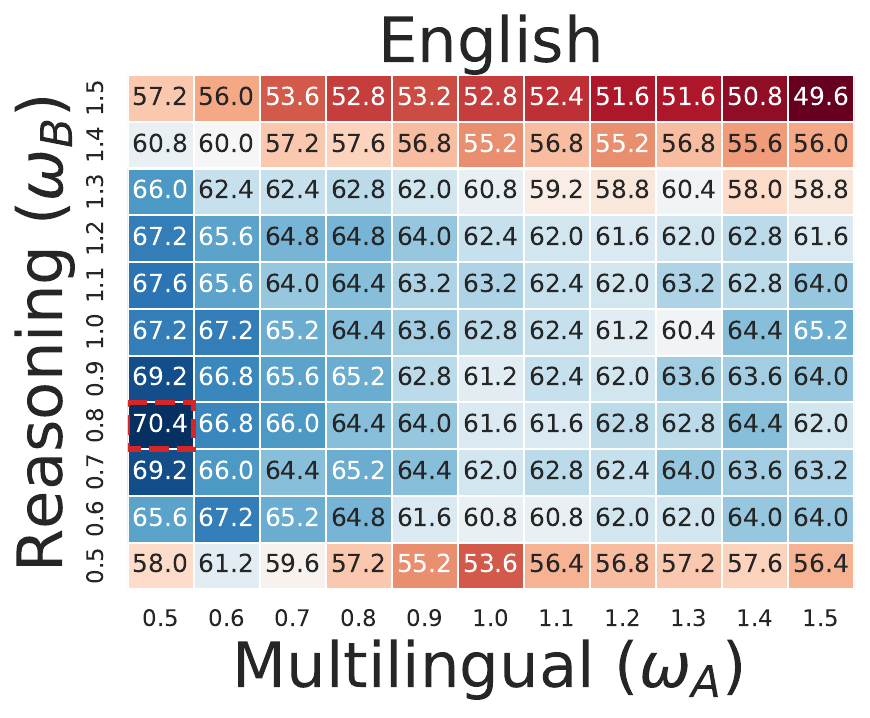} &
    \hspace{-0.35cm} \includegraphics[width=0.32\linewidth]{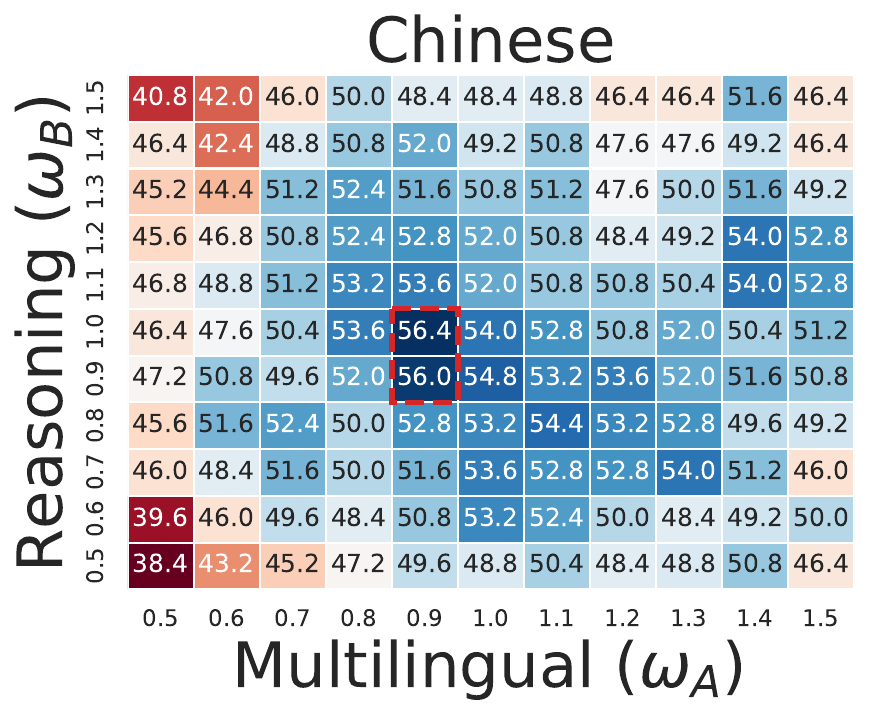} & 
    \hspace{-0.35cm} \includegraphics[width=0.32\linewidth]{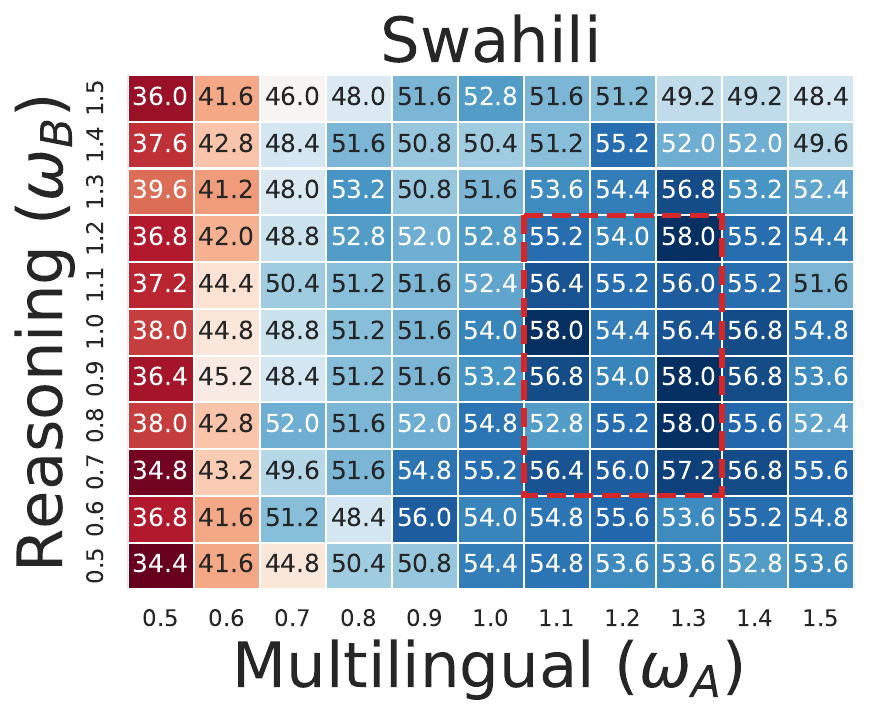}\\
\end{tabular}
\caption{Accuracy on MGSM with different manual weight combinations for the two source models. Darker blue grids indicate higher reasoning accuracy.}
\label{fig:manual_weight}
\end{figure}

In this paper, we first conduct an exploratory analysis by introducing manual scalar weights, $\omega_A$ and $\omega_B$, to modulate the representation intensity of the multilingual encoder and the reasoning LLM, respectively. As illustrated in Figure~\ref{fig:manual_weight}, the heatmap of accuracy  exhibits divergent collaboration patterns across languages. For English, optimal performance is achieved when the weight for the multilingual encoder  ($\omega_A$) is relatively low. This suggests that for languages where the base LLM is already proficient, over-reliance on external multilingual signals may act as noise interfering  with the inherent reasoning pathways of LLM. Conversely, for Swahili, performance peaks only when the multilingual encoder contributes significantly (high $\omega_A$). Here, the base LLM lacks the necessary linguistic grounding, and amplifying the external representation is crucial to bridge the semantic gap and activate the reasoning capabilities.  

To pursue both effective utilization of low-resource language understanding and the preservation of inherent reasoning abilities, we propose a Steerable Model Merging framework (\textbf{ST-Merge}) with gated cross-attention. Instead of relying on a fixed concatenation of representations, our method enables the model to dynamically modulate the contribution of each source model (i.e., the multilingual encoder and the reasoning LLM), allowing for more flexible and adaptive coordination. This design facilitates input-aware modulation modeling, enabling the merged model to shift its inductive bias toward the source most aligned with the current input. As a result, it yields more accurate and targeted reasoning across diverse linguistic contexts. The main contributions of this paper are as follows:

\begin{itemize}
    \item We propose a steerable model merging (\textbf{ST-Merge}) framework that modulates multilingual understanding and reasoning preservation for multilingual reasoning.
    \item We devise a gated cross-attention mechanism to dynamically modulate the contributions of source models.
    \item Extensive experiments on four multilingual reasoning benchmarks across 21 languages demonstrate that ST-Merge consistently outperforms strong baselines.
\end{itemize}

\section{Related Work}
\label{sec:relatedwork}
\subsection{Multilingual Reasoning}
Enhancing multilingual reasoning in English-centric LLMs remains a critical challenge. Existing approaches can be broadly categorized into translation-based methods and model merging paradigms. Translation-based strategies, which involve fine-tuning on translated datasets~\cite{zhu-etal-2024-question/qalign} or employing external translators~\cite{DBLP:conf/iclr/ShiSF0SVCTRZ0W23/mgsm}, have demonstrated significant performance gains. However, these methods incur substantial computational overheads due to the heavy reliance on high-quality parallel corpora and the computational latency of autoregressive decoding. Conversely, model merging has recently emerged as a popular alternative~\cite{yoon-etal-2024-langbridge, NEURIPS2024_3bf80b34/mindmerger}, aiming to combine the strengths of different experts. Despite their promise, current merging techniques typically employ static fusion strategies, often overlooking the inherent feature conflicts and interference between the multilingual encoder and the reasoning LLM. To address this, we propose ST-Merge, which introduces a dynamic gated network. This approach allows for the flexible, context-aware merging of models, effectively resolving inter-model conflicts while maximizing the collaboration between multilingual understanding and logical reasoning.

\subsection{Model Merging}

Model merging aims to combine the strengths of multiple models into a unified architecture and has been widely used to enhance capabilities such as modality integration~\citep{DBLP:conf/emnlp/SungLLGBW23/Multimodal2,DBLP:conf/acl/0005DFWLLYZH0024/Multimodal} and task generalization~\citep{DBLP:conf/iclr/BandarkarMYHSLL25/layerswap,DBLP:conf/cvpr/DuWCYWLYZHS0L25/adamms}. Existing works can be broadly categorized into two types: homogeneous merging, which combines models with the same architecture, and heterogeneous merging, which merges models across architectural or modality boundaries. Recent studies have explored model merging for cross-lingual transfer learning~\citep{yoon-etal-2024-langbridge,NEURIPS2024_3bf80b34/mindmerger}, but often suffer from limited controllability and alignment issues in multilingual reasoning settings. In contrast, our work introduces a steerable model merging approach that dynamically modulates the representations, enabling better coordination between multilingual encoder and the reasoning LLM for reasoning across both low-resource and high-resource languages.

\subsection{Gated Attention Mechanism}

Gated cross-attention mechanisms have been developed to selectively fuse heterogeneous representations by leveraging learnable weights~\cite{DBLP:conf/aaai/ChaplotSPRS18,DBLP:conf/wacv/LeeYJ22}. These approaches employ multiplicative or residual gating strategies to dynamically weight features, effectively filtering noise and enhancing interpretability in  fusion tasks~\cite{DBLP:conf/mlhc/KimS21,DBLP:journals/kbs/OrtizPerezBRRT25}. Recent studies have extended this paradigm to various architectures, including router-based gating for audio-visual recognition and sparse gating for large language models~\cite{DBLP:conf/cvpr/JeongPKK25,DBLP:journals/corr/abs-2505-06708}. However, while these mechanisms have proven effective in multimodal settings, their potential for steering cross-lingual alignment within a model merging framework remains unexplored.

\section{Steerable Model Merging}

\begin{figure}[t]
\begin{center}
\includegraphics[width=0.45\textwidth]{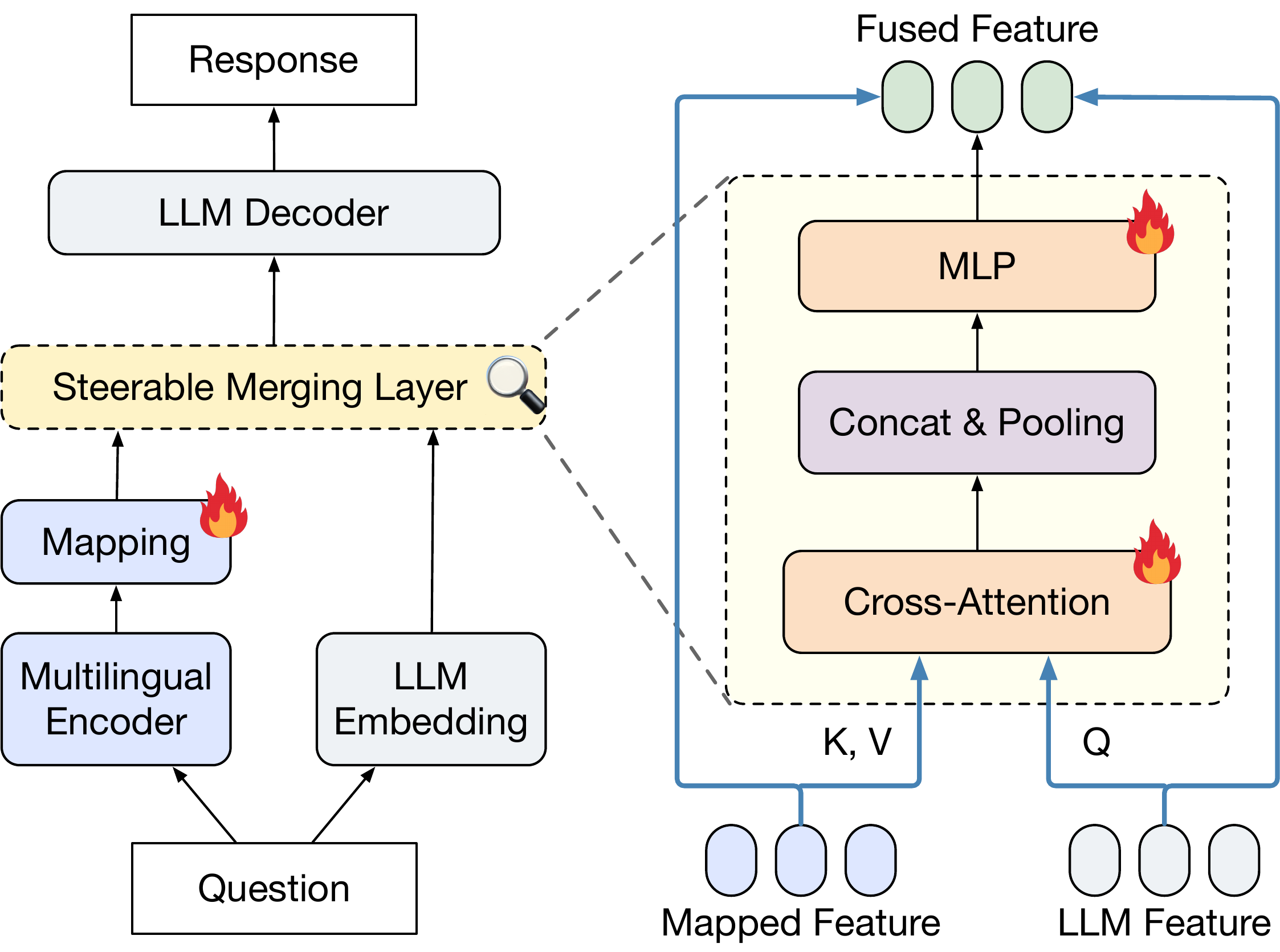}
\end{center}
\caption{
Framework of the proposed steerable model merging \textbf{ST-Merge} method for multilingual reasoning. 
}
\label{fig:framework}
\end{figure}

In this section, we introduce our Steerable Model {Merging} (\textbf{ST-Merge}) method, which is designed to weight and filter attended models conditioned on the specific input question. Figure \ref{fig:framework} depicts an overview of the ST-Merge framework. We will provide a detailed description of our approach from the following two stages: feature space alignment and gated cross-attention learning.

\paragraph{Problem Formulation}
Multilingual reasoning can be formulated as a text generation task. Given an input sequence $\mathbf{x}$ (e.g. a math problem), the model aims to generate the target output sequence $\mathbf{y}$ (e.g. a chain-of-thought and the answer). Formally, the language modeling likelihood of the target output is denoted
as: \begin{equation}
p(\mathbf{y}|\mathbf{x}) = \prod_{i}^{L} p(y_i|\mathbf{x}, y_{<i})
\end{equation} Under the paradigm of model merging, we assume there are a multilingual encoder $\mathbf{m}_A$ and a reasoning LLM $\mathbf{m}_B$. Our goal is to learn a merger $\mathbf{m}_{A \oplus B}$  that optimizes the generation probability $p(\mathbf{y} | \mathbf{x})$, ensuring the reasoning accuracy is maintained across different languages. 

\subsection{Feature Space Alignment}
First, we extract distinct features for each input sequence. Specifically, we utilize a multilingual encoder to capture linguistic understanding features and a Large Language Model (LLM) to extract reasoning features. 
\paragraph{Multilingual Feature} Multilingual feature extraction is performed by the mT5 encoder~\cite{xue-etal-2021-mt5}. Given an input sequence $\mathbf{x}$, we employ the multilingual model to encode it into a generalized representation $\mathbf{H}_{A}$, thereby mitigating the complexity of cross-lingual understanding:
\begin{equation}
\mathbf{H}_{A} = \text{Encoder}(\mathbf{x})
\end{equation}
where $\mathbf{H}_{A} \in \mathbb{R}^{l_A \times d_A}$ is the hidden state output of the last layer in the multilingual encoder, with $l_A$ denoting the sequence length of the original input and $d_A$ the hidden dimension.

\paragraph{Reasoning Feature} 
We utilize LLaMA-based parameter-efficient fine-tuned reasoning LLM (e.g. MetaMath~\cite{DBLP:conf/iclr/YuJSYLZKLWL24/metamath}) to extract reasoning features. To fully activate the intrinsic reasoning capabilities of the LLM, we append a chain-of-thought prompt $\mathbf{p}$ (e.g., ``Let's think step by step'') to the original input $\mathbf{x}$, forming a prompted sequence $\mathbf{x}' = [\mathbf{x}; \mathbf{p}]$. We then process this sequence directly through the embedding layer:
\begin{equation}
\mathbf{H}_{B} = \text{Embedding}(\mathbf{x}')
\end{equation}
where $\mathbf{H}_{B} \in \mathbb{R}^{l_B \times d_B}$ represents the representation within the semantic space of the reasoning LLM, with $l_B$ denoting the sequence length of the original input with the prompt and $d_B$ the LLM embedding dimension.

\paragraph{Feature Alignment}
Since the representation $\mathbf{H}_{A}$ resides in the multilingual representation space, which is separate from the reasoning LLM space. The extracted features cannot be used for reasoning directly. Therefore, we project the multilingual features via a mapping layer:

\begin{equation}
\mathbf{\hat{H}}_A = \text{Mapping}(\mathbf{H}_A)
\end{equation}
where $\mathbf{\hat{H}}_A \in \mathbb{R}^{l_A \times d_B}$ is the projection of $\mathbf{H}_{A}$ onto the reasoning feature space. Unless otherwise stated, $\text{Mapping}(\cdot)$ is implemented as a two-layer Multi-Layer Perceptron (MLP). This transformation aligns the semantic spaces of the multilingual encoder and the reasoning LLM, enabling effective feature merging despite the frozen weights of the base models.

\subsection{Gated Cross-Attention Learning}

To overcome the limitations of fixed model merging, we introduce a gated cross-attention network to dynamically estimate the optimal weights for the multilingual features $\mathbf{\hat{H}}_A$ and the reasoning features $\mathbf{H}_B$ conditioned on the specific input. 

\paragraph{Cross-Attention}  We first facilitate a comprehensive information interaction between the two types of features to construct a holistic context for  gating estimation. Formally, the reasoning feature $\mathbf{H}_B$ acts as the query ($\mathbf{Q}_B$) to attend to the aligned multilingual representation $\mathbf{\hat{H}}_A$, which serves as the key ($\mathbf{K}_A$) and value ($\mathbf{V}_A$). We employ $\mathbf{H}_B$ as the query to ensure the attention mechanism is anchored in the semantic space of the reasoning task. 
\begin{gather}
    \mathbf{K}_{A}, \mathbf{V}_{A} = \mathbf{\hat{H}}_{A}\mathbf{W}_{k}^{K}, ~ \mathbf{\hat{H}}_{A}\mathbf{W}_{k}^{V} \\
    \mathbf{Q}_{B} = \mathbf{H}_{B}\mathbf{W}_{k}^{Q} \\
    \text{head}_{k} = \text{Attn.}(\mathbf{Q}_{B}, \mathbf{K}_{A}, \mathbf{V}_{A}) \\
    \mathbf{G}_{A \oplus B} = \text{Concat.}_{k} (\text{head}_{k}) \mathbf{W}^{O}
\end{gather}
where $\mathbf{W}_{k}^{Q}, \mathbf{W}_{k}^{K}, \mathbf{W}_{k}^{V} \in \mathbb{R}^{d_{B} \times d_k}$ denote the projection matrices for the $k$-th head, and $d_k$ is the dimension of each attention head. $\mathbf{W}^{O} \in \mathbb{R}^{d_{B} \times d_{B}}$ is the output projection matrix used to aggregate information from all $k$ heads. 

\paragraph{Language Embedding} 
We introduce a learnable lightweight language embedding to explicitly inject language identity, which facilitates language differentiation. Given the language ID, we retrieve the corresponding embedding vector $\mathbf{E}_{Lang} \in \mathbb{R}^{d_L}$ and concatenate it with the global context $\mathbf{G}_{A \oplus B} \in \mathbb{R}^{d_B}$, yielding a composite representation $\mathbf{Z} \in \mathbb{R}^{d_B + d_L}$.

\paragraph{Feature Fusion} 
Finally, we employ an MLP layer to project $\mathbf{Z}$ to generate the two weights to modulate the $\mathbf{\hat{H}}_A$ and $\mathbf{H}_B$. We specifically utilize a $1 + \tanh$ activation function to center the weights around $1$, as the value of $1$ indicates the fixed model merging with original features providing a stable initialization. The final input to the LLM decoder is constructed as follows:
\begin{gather}
    \hspace*{-2mm}
    [\mathbf{\omega}_A, \mathbf{\omega}_B] = 1 + \tanh(\text{MLP}(\mathbf{Z}))\\
    \hspace*{-1mm}
    \mathbf{H}_{A \oplus B} = [ \langle \text{bos} \rangle ; \mathbf{\omega}_A \cdot \mathbf{\hat{H}}_{A} ; \langle \text{sep} \rangle ; \mathbf{\omega}_B \cdot \mathbf{H}_{B} ]
\end{gather}
where $\mathbf{\omega}_A, \mathbf{\omega}_B$ are two input-dependent scalar weights for the multilingual and reasoning representations, respectively; $\langle \text{bos} \rangle$ and $\langle \text{sep} \rangle$ are learnable boundary tokens. The resulting fused embedding $\mathbf{H}_{A \oplus B}$ serves as the steered input to the frozen LLM, guiding the generation of the chain-of-thought reasoning path and the final response.

\section{Experiments}
\label{sec:experiments}

\begin{table*}[t]
\setlength{\tabcolsep}{4pt}
\label{tab:main-mgsm}
\begin{center}
\begin{tabular}{l|ccc|ccccccc|ccc}
\toprule
\textbf{MGSM} & \textbf{Bn} & \textbf{Th} & \textbf{Sw} & \textbf{Ja} & \textbf{Zh} & \textbf{De} & \textbf{Fr} & \textbf{Ru} & \textbf{Es} & \textbf{En} & \textbf{Lrl.} & \textbf{Hrl.} & \textbf{Avg.} \\
 \hline

Translate-En~[\citeyear{DBLP:conf/iclr/ShiSF0SVCTRZ0W23/mgsm}] & 48.4 & 37.6 & 37.6 & 49.2 & 46.8 & 60.4 & 56.4 & 47.6 & 59.6 & 65.5 & 41.2 & 55.1 & 50.6 \\
MetaMath~[\citeyear{DBLP:conf/iclr/YuJSYLZKLWL24/metamath}] & 6.8 & 7.2 & 6.8 & 36.4 & 38.4 & 55.2 & 54.4 & 52.0 & 57.2 & \textbf{68.8} & 6.9 & 51.8 & 38.3 \\
MultiReason~[\citeyear{zhu-etal-2024-question/qalign}] & 33.2 & 40.0 & 42.0 & 42.0 & 42.0 & 45.2 & 44.8 & 45.2 & 48.0 & 52.0 & 38.4 & 45.6 & 43.4 \\
QAlign~[\citeyear{zhu-etal-2024-question/qalign}] & 32.4 & 39.6 & 40.4 & 44.0 & 48.4 & 54.8 & 56.8 & 52.4 & 59.6 & 68.0 & 37.5 & 54.9 & 49.6 \\
LangBridge~[\citeyear{yoon-etal-2024-langbridge}] & 42.8 & 50.4 & 43.2 & 40.0 & 45.2 & 56.4 & 50.8 & 52.4 & 58.0 & 63.2 & 45.5 & 52.3 & 50.2 \\
MindMerger~[\citeyear{NEURIPS2024_3bf80b34/mindmerger}] & 50.4 & 52.8 & 57.2 & \textbf{54.4} & 53.6 & 61.2 & 57.6 & 60.8 & 58.4 & 66.8 & 53.5 & 59.0 & 57.3 \\
LayAlign~[\citeyear{ruan-etal-2025-layalign}] & 51.6 & \textbf{59.2} & 58.4 & 52.0 & 56.0 & 62.0 & \textbf{61.6} & 61.6 & {61.6} & 66.4 & 56.4 & 60.2 & 59.0 \\

 \rowcolor{lightgray!20}

\textbf{$\text{ST-Merge (Ours)}$} & \textbf{54.0} & {56.8} & \textbf{58.8} & {53.5} & \textbf{57.2} & \textbf{62.4} & {61.2} & \textbf{62.8} & \textbf{65.2} & {68.0} & \textbf{56.5} & \textbf{61.5} & \textbf{60.0} \\

\bottomrule
\end{tabular}
\setlength{\tabcolsep}{4pt}
\begin{tabular}{l|ccc|ccccccc|ccc}
\toprule
\textbf{MSVAMP} & \textbf{Bn} & \textbf{Th} & \textbf{Sw} & \textbf{Ja} & \textbf{Zh} & \textbf{De} & \textbf{Fr} & \textbf{Ru} & \textbf{Es} & \textbf{En} & \textbf{Lrl.} & \textbf{Hrl.} & \textbf{Avg.} \\
 \hline

Translate-En~[\citeyear{DBLP:conf/iclr/ShiSF0SVCTRZ0W23/mgsm}] & 47.9 & 51.3 & 43.1 & 50.4 & 55.8 & 43.9 & 50.9 & 53.4 & 51.4 & 60.6 & 47.4 & 52.3 & 50.9 \\
MetaMath~[\citeyear{DBLP:conf/iclr/YuJSYLZKLWL24/metamath}]  & 14.4 & 19.5 & 16.8 & 53.4 & 55.0 & 63.5 & 64.1 & 60.3 & 64.9 & 66.3 & 16.9 & 61.1 & 47.8 \\
MultiReason~[\citeyear{zhu-etal-2024-question/qalign}] & 34.8 & 38.1 & 39.8 & 43.4 & 42.9 & 45.6 & 45.8 & 45.0 & 46.1 & 46.8 & 37.6 & 45.1 & 42.8 \\
QAlign~[\citeyear{zhu-etal-2024-question/qalign}] & 41.7 & 47.7 & 54.8 & 58.0 & 55.7 & 62.8 & 63.2 & 61.1 & 63.3 & 65.3 & 48.1 & 61.3 & 57.2 \\
LangBridge~[\citeyear{yoon-etal-2024-langbridge}] & 46.8 & 46.3 & 42.1 & 45.5 & 50.4 & 58.1 & 57.0 & 55.8 & 56.9 & 60.6 & 45.1 & 54.9 & 52.0 \\
MindMerger~[\citeyear{NEURIPS2024_3bf80b34/mindmerger}] & 52.0 & 53.4 & 54.0 & 59.0 & \textbf{61.7} & \textbf{64.1} & 64.0 & \textbf{63.3} & 65.0 & \textbf{67.7} & 53.1 & 63.5 & 60.4 \\
LayAlign~[\citeyear{ruan-etal-2025-layalign}]  & 51.8 & 55.1 & 56.9 & 59.3 & 58.7 & 62.5 & 62.1 & 58.8 & 62.0 & 64.0 & 54.6 & 61.1 & 59.1\\
 \rowcolor{lightgray!20}
\textbf{$\text{ST-Merge (Ours)}$} & \textbf{52.8} & \textbf{56.3} & \textbf{57.6} & \textbf{59.7} & \textbf{61.7} & 63.8 & \textbf{65.7} & 62.2 & \textbf{66.1} & 67.6 & \textbf{55.6} & \textbf{63.9} & \textbf{61.4 }\\
\bottomrule
\end{tabular}
\caption{
Accuracy (\%) results on MGSM and MSVAMP. We regard Bn, Th, and Sw as low-resource languages, and regard the remaining languages as high-resource languages. Lrl., Hrl., and Avg. represent the average accuracy across low-resource languages, high-resource languages, and all languages, respectively. The best performance is in bold (same for Table \ref{tab:main-xcsqa} and Table \ref{tab:main-xnli}).
}
\label{tab:main-mgsm}
\end{center}
\end{table*}

\subsection{Evaluation Datasets} 
We evaluate models on four multilingual reasoning datasets across 21 different languages: 

\paragraph{Mathematical Reasoning}  We evaluate on the multilingual math problem datasets MGSM and MSVAMP for this task. \textbf{MGSM}~\citep{DBLP:conf/iclr/ShiSF0SVCTRZ0W23/mgsm} consists of grade-school level math questions translated by humans into 11 typologically diverse languages. \textbf{MSVAMP}~\citep{DBLP:conf/emnlp/0001ZWG0024/MSVAMP} extends the SVAMP dataset~\citep{patel-etal-2021-nlp/svamp} to 10 languages, offering linguistically diverse paraphrases of math problems with varying reasoning structures. 

\paragraph{Commonsense Reasoning}  We evaluate commonsense reasoning using \textbf{X-CSQA}~\citep{lin-etal-2021-common/xcsqa}, a multilingual extension of the CommonsenseQA dataset. X-CSQA provides translated versions of CSQA across multiple languages, along with a new data split to support cross-lingual evaluation. The dataset includes 8,888 English training examples, 1,000 development examples per language, and 1,074 test examples per language. 

\begin{table}[t]
\setlength{\tabcolsep}{3pt}
\begin{center}
\begin{tabular}{l|ccc|c}
\toprule
\textbf{X-CSQA} & \textbf{Sw} & \textbf{Fr} & \textbf{En} & \textbf{Avg.} \\
 \hline
Translate-En~[\citeyear{DBLP:conf/iclr/ShiSF0SVCTRZ0W23/mgsm}] & 36.5 & 57.2 & 71.3 & 52.3 \\
MetaMath~[\citeyear{DBLP:conf/iclr/YuJSYLZKLWL24/metamath}] & 24.2 & 63.5 &  76.3 & 51.3\\
MultiReason~[\citeyear{zhu-etal-2024-question/qalign}] & 27.6 & 52.1 & 67.2 & 43.8\\
QAlign~[\citeyear{zhu-etal-2024-question/qalign}] & 35.1 & 60.3 & 75.7 & 52.3 \\
LangBridge~[\citeyear{yoon-etal-2024-langbridge}] & 31.8 & 38.2 & 44.4 & 36.1 \\
MindMerger~[\citeyear{NEURIPS2024_3bf80b34/mindmerger}] & 45.5  & \textbf{68.1} & \textbf{78.1} & 61.0 \\
LayAlign~[\citeyear{ruan-etal-2025-layalign}] & 53.3 & 66.5 & 76.7 & 62.3\\
 \rowcolor{lightgray!20}
\textbf{$\text{ST-Merge (Ours)}$} & \textbf{53.6} & 67.2 & 77.3 & \textbf{62.5}\\
\bottomrule
\end{tabular}
\end{center}
\caption{Accuracy (\%) on X-CSQA. Avg. represents the average accuracy across all languages.}
\label{tab:main-xcsqa}
\end{table}

\paragraph{Natural Language Inference} We evaluate natural language inference using \textbf{XNLI}~\citep{conneau-etal-2018-xnli}, a widely used multilingual benchmark spanning 15 languages. The task involves determining whether a given \textit{hypothesis} logically follows from a \textit{premise}, categorized as entailment, contradiction, or neutral. The dataset covers languages both typologically close to English (e.g., French, German, Spanish) and more distant (e.g., Arabic, Thai, Swahili), making it well-suited for evaluating cross-lingual generalization.

\begin{table}[t]
\setlength{\tabcolsep}{3pt}
\begin{center}
\begin{tabular}{ l | ccc | c }
\toprule
\textbf{XNLI} & \textbf{Sw} & \textbf{Fr} & \textbf{En} & \textbf{Avg.} \\
\hline
Translate-En~[\citeyear{DBLP:conf/iclr/ShiSF0SVCTRZ0W23/mgsm}] & 65.3 & 80.4 & 81.4 & 75.1 \\
MetaMath~[\citeyear{DBLP:conf/iclr/YuJSYLZKLWL24/metamath}]  & 45.9 & 82.2 & \textbf{90.0} & 68.7\\
MultiReason~[\citeyear{zhu-etal-2024-question/qalign}]  & 56.3 & 82.9 & 88.8 & 71.9\\
QAlign~[\citeyear{zhu-etal-2024-question/qalign}] & 65.2  & 83.1 & 89.1 & 73.5\\
LangBridge~[\citeyear{yoon-etal-2024-langbridge}] & 71.7 & 79.9 & 83.4 & 76.5 \\
MindMerger~[\citeyear{NEURIPS2024_3bf80b34/mindmerger}] & 66.6 & 83.9 & 88.7 & 78.4 \\
LayAlign~[\citeyear{ruan-etal-2025-layalign}] & 73.0 & 84.7 & 88.9 & 79.7 \\
 \rowcolor{lightgray!20}

\textbf{$\text{ST-Merge (Ours)}$} & \textbf{73.7}  & \textbf{84.8} & {89.1} & \textbf{79.9} \\
\bottomrule
\end{tabular}
\end{center}
\caption{Accuracy (\%) on XNLI. Avg. represents the average accuracy across all languages.}
\label{tab:main-xnli}
\end{table}

\begin{table*}[!t]
\setlength{\tabcolsep}{3pt}
\begin{center}
\begin{tabular}{l|ccc|ccccccc|ccc}
\toprule
\textbf{MGSM} & \textbf{Bn} & \textbf{Th} & \textbf{Sw} & \textbf{Ja} & \textbf{Zh} & \textbf{De} & \textbf{Fr} & \textbf{Ru} & \textbf{Es} & \textbf{En} & \textbf{Lrl.} & \textbf{Hrl.} & \textbf{Avg.} \\
\midrule
\rowcolor{lightgray!20}
\textbf{$\text{ST-Merge (Ours)}$} & \textbf{54.0} & \textbf{56.8} & \textbf{58.8} & \textbf{53.5} & \textbf{57.2} & \textbf{62.4} & \textbf{61.2} & \textbf{62.8} & \textbf{65.2} & \textbf{68.0} & \textbf{56.5} & \textbf{61.5} & \textbf{60.0} \\
\textit{\quad w/o Lang. Embed} & 51.4 & 54.0 & 56.4 & 53.2 & 56.6 & {61.4} & 60.3 & 62.1 & {63.9} & 67.6 & 53.9 & 60.7 & 58.7\\
\textit{\quad w/o Cross-Attention} & 52.7 & 54.7 & 57.8 & 52.8 & 55.8 & 60.8 & 58.5 & {62.7} & 63.4 & 67.1 & 55.1 & 60.2 & 58.6 \\
\textit{\quad w/o Gate Network} & 50.7 & 52.4 & 55.8 & 51.0 & 53.6 & 59.8 & 56.4 & 60.7 & 62.0 & 66.5 & 53.0 & 58.6 & 56.9 \\
{Fix-Merge (Baseline)} & 50.5 & 52.9 & 55.7 & 50.8 & 54.8 & 59.1 & 56.8 & 60.7 & 61.7 & 66.7 & 53.0 & 58.7 & 57.0 \\
\bottomrule
\end{tabular}
\end{center}
\caption{
Ablation study on MGSM.}
\label{tab:ablation}
\end{table*}

\subsection{Implementation Details} 
Following prior setup~\citep{NEURIPS2024_3bf80b34/mindmerger,ruan-etal-2025-layalign}, we train the mapping layer using the Lego-MT corpus~\cite{yuan-etal-2023-lego} via translation tasks. Subsequently, we leverage the MultilingualMath dataset~\cite{DBLP:conf/iclr/YuJSYLZKLWL24/metamath,DBLP:conf/emnlp/0001ZWG0024/MSVAMP} for the gated cross-attention network learning. We adopt the encoder of mT5-xl~\citep{xue-etal-2021-mt5} as the multilingual backbone, and employ MetaMath~\citep{DBLP:conf/iclr/YuJSYLZKLWL24/metamath} as the large language reasoning model across all experiments, ensuring a fair comparison with prior work~\citep{yoon-etal-2024-langbridge,NEURIPS2024_3bf80b34/mindmerger,ruan-etal-2025-layalign}. The final model is selected based on the averaged performance of all languages on the dev set. For training, we utilized 4 NVIDIA A100 GPUs with a learning rate of 2e-5, a batch size of 128, a maximum sequence length of 512, and a total of 3 epochs. We conduct experiments with three different random seeds and report the average results. 

\subsection{Baselines} 
We compare our method against several state-of-the-art baselines for multilingual reasoning: \\
\textbf{Translate-En}~\citep{DBLP:conf/iclr/ShiSF0SVCTRZ0W23/mgsm} translates non-English inputs to English and uses an English reasoning model. \\
\textbf{MetaMath}~\citep{DBLP:conf/iclr/YuJSYLZKLWL24/metamath} is fine-tuned from LLaMA2-7B on an additional mathematical dataset MetaMathQA, which serves as the backbone architectures for baseline methods. \\
\textbf{MultiReason}~\citep{zhu-etal-2024-question/qalign} enhances reasoning consistency across languages via question alignment and rationale generation. \\
\textbf{QAlign}~\citep{zhu-etal-2024-question/qalign} aligns questions across languages through fine-tuned translation-based contrastive learning. \\
\textbf{LangBridge}~\citep{yoon-etal-2024-langbridge} introduces an alignment layer to bridge non-English inputs to an English-centric reasoning space. \\
\textbf{MindMerger}~\citep{NEURIPS2024_3bf80b34/mindmerger} merges task representations across languages to promote cross-lingual reasoning alignment. \\
\textbf{LayAlign}~\citep{ruan-etal-2025-layalign} integrates representations from all encoder layers to enable layer-wise interaction between the two models.

\subsection{Main Results} 

As shown in Table~\ref{tab:main-mgsm}, Table~\ref{tab:main-xcsqa}, and Table~\ref{tab:main-xnli}, {ST-Merge} achieves consistent performance gains across all tasks, demonstrating strong generalization. (Please refer to Appendix Table~\ref{app_tab:main-xcsqa}, Table~\ref{app_tab:main-xnli} for the complete results of X-CSQA and XNLI.) The performance gains on the X-CSQA are relatively limited. We conjecture that the limited gains stem from their discrete choice-format outputs, which provide limited signals and hinder effective optimization of the gate network.

Compared to existing strong multilingual reasoning baselines such as MindMerger~\cite{NEURIPS2024_3bf80b34/mindmerger}, our method achieves superior performance under identical prompts and training data. Our approach can act as a plug-and-play enhancement over fixed model merging strategies, yielding average gains of +1.7\%, +1.3\%, +1.5\%, and +1.5\% over the MindMerger baseline on the four benchmarks, respectively. 

Notably, our method consistently improves average performance on high-resource languages. As shown in Table~\ref{tab:main-mgsm}, compared to the state-of-the-art baseline, {ST-Merge} achieves an average  high-resource gain +1.3\% on MGSM and +0.4\% on MSVAMP, respectively. ST-Merge maintains highly competitive accuracy in English (68.0\% on MGSM), outperforming other multilingual baselines while effectively narrowing the gap with the state-of-the-art. This demonstrates that ST-Merge can preserve reasoning proficiency in high-resource languages.

\begin{table*}[t]
\centering
\small
\renewcommand{\arraystretch}{1.3}
\resizebox{\textwidth}{!}{
\begin{CJK*}{UTF8}{gbsn} %
\begin{tabular}{p{4.5cm} p{5.7cm} p{5.7cm}}
\toprule
\textbf{Question [Swahili]:} John hukimbia maili 60 kwa wiki. Huwa anakimbia siku 3 kwa wiki. Huwa anakimbia saa 3 siku ya kwanza na nusu ya kiasi hicho kwa siku nyingine mbili anazokimbia. Huwa anakimbia kwa kasi gani?  \quad (\textbf{Answer:} 10) & 
\textbf{Output [Fix-Merge]:} John runs 3 miles on the first day. He runs half that distance on the other two days, so he runs 3/2 = 1.5 miles on each of those days. In total, he runs 3 + 1.5 + 1.5 = 6 miles in a week. Since he runs 60 miles in a week, his speed is \loss{6\underline{\textbf{ }}$_{[\text{ \xmark}, 0.5523]}$miles / 7 days = 0.857} miles per day. The answer is: 0.857. & 
\textbf{Output [ST-Merge (Ours)]:} John runs 3 miles on the first day. He runs half that distance on the other two days, so he runs 3/2 = 1.5 miles on each of those days. In total, he runs 3 + 1.5 + 1.5 = 6 miles in a week. Since he runs 60 miles in a week, his speed is \win{6\underline{\textbf{0}}$_{[\text{ \cmark},0.9045]}$/6 = 10} miles per hour. The answer is: 10.  \quad \underline{({$\omega_A$: 1.1172, $\omega_B$: 1.0547})} \\
\cmidrule(lr){1-3}

\textbf{Question [French]:} Kylar se rend au magasin afin d'acheter des verres pour son nouvel appartement. Un verre coûte 5 \$, mais chaque deuxième verre ne coûte que 60\% du prix. Kylar veut acheter 16 verres. Combien devra-t-il payer pour les acheter? \quad (\textbf{Answer:} 64)& 

\textbf{Output [Fix-Merge]:} The first glass costs \$5. The second glass costs 60\% of \$5, which is 0.6 * \$5 = \$3. So for every two glasses, Kylar pays \$5 + \$3 = \$8. Kylar wants to buy 16 glasses, so he will pay \loss{16/2 * \$8 = \$\underline{\textbf{8}}$_{[\text{ \xmark}, 0.4948]}$0 }. The answer is: 80. & 

\textbf{Output [ST-Merge (Ours)]:} The first glass costs \$5. The second glass costs 60\% of \$5, which is 0.6 * \$5 = \$3. For every two glasses, the price is \$5 + \$3 = \$8. Kylar wants to buy 16 glasses, so he will need to pay \win{16/2 * \$8 = \$\underline{\textbf{6}}$_{[\text{ \cmark}, 0.9614]}$4}. The answer is: 64. \quad \underline{({$\omega_A$: 1.0156, $\omega_B$: 1.0503})}\\

\cmidrule(lr){1-3}

\textbf{Question [Chinese]:} 帕姆和弗莱德去参加狂欢节。帕姆坐了 2 次云霄飞车，而弗莱德坐了 4 次。在此之后，他们各自决定坐雪橇 2 次。如果每坐一次需要 6 张门票，他们那天用掉了多少张门票？ \quad (\textbf{Answer:} 60) & 
\textbf{Output [Fix-Merge]:} Pam rides the roller coaster 2 times, so she uses 2 x 6 = 12 tickets. Fred rides the roller coaster 4 times, so he uses 4 x 6 = 24 tickets. After riding the roller coaster, they each decide to ride the ice cream, so they each use 2 x 6 = 12 tickets. In total, they use \loss{12 + \underline{\textbf{1}}$_{[\text{ \xmark}, 0.7773]}$2 + 24 + 24 = 72} tickets. The answer is: 72. & 
\textbf{Output [ST-Merge (Ours)]:} Pam rides the roller coaster 2 times, so she uses 2 x 6 = 12 tickets. Fred rides the roller coaster 4 times, so he uses 4 x 6 = 24 tickets. After riding the roller coaster, they each decide to ride the snow sled 2 times, so they use 2 x 6 = 12 tickets each. In total, they use 12 + 12 = 24 tickets for the snow sled. Therefore, they use a total of \win{12 + \underline{\textbf{2}}$_{[\text{ \cmark}, 0.9993]}$4 + 24 = 60} tickets that day. The answer is: 60. \underline{({$\omega_A$: 0.9766, $\omega_B$: 0.9314})}\\

\bottomrule
\end{tabular}
\end{CJK*}
}
\caption{Case study on MGSM. The GREEN (RED) highlight indicates a correct (incorrect) reasoning step. The real-valued numbers indicate the next token generation probability. ($\omega_A$, $\omega_B$) represent the learned weights by our method for each language.}
\label{tab:case_study} 
\end{table*}

\subsection{Ablation Study}

Table~\ref{tab:ablation} presents the ablation results on the MGSM.

\begin{enumerate}

\item[(1)]  \textit{w/o Lang. Embed}, which removes the language identity embeddings from the gating network. The average accuracy drops from 60.0\% to 58.7\%, with a notable degradation of 2.6\% on low-resource languages. This suggests that without explicit language cues, the gating network lacks the guidance to differentiate between languages. Consequently, the optimization process becomes biased towards dominant high-resource languages, hindering the low-resource language reasoning capacity of the merger model.

\item[(2)]  \textit{w/o Cross-Attention}, which replaces the fine-grained Cross-Attention mechanism with a simpler concatenation of the representations. Removing this module leads to a significant performance decline across all languages, reducing the average accuracy to 58.6\%. This suggests that the token-level interaction provided by cross-attention is essential for deeply considering the context from the two models.

\item[(3)]  \textit{w/o Gate Network}, which completely eliminates the gating network. In this case, we increase the number of training steps to match the computational budget of the proposed ST-Merge. Despite the extended training, this variant shows no improvement over the static baseline (Fix-Merge, 57.0\%) and lags significantly behind the proposed full ST-Merge model (60.0\%). This validates the indispensability of the gated cross-attention, demonstrating that the performance boost is driven by the steerable merging strategy rather than simply extending the optimization process.

\end{enumerate}

\subsection{Case Study}
We present a case study to show that the failed cases of Fix-Merge (fixed model merging strategy) can be rectified by our model. We aim to provide insights into the mechanisms underlying the effectiveness of the proposed steerable model merger.

The proposed ST-Merge framework adaptively modulates feature amplification and suppression to extract the most effective representations for multilingual reasoning. Specifically, ST-Merge leverages the complementary strengths of the backbone LLM, which possesses strong intrinsic reasoning capabilities, and the external multilingual model, which excels in cross-lingual semantic understanding. By utilizing learned weights to selectively enhance these respective strengths while suppressing irrelevant noise, the model is steered toward more accurate reasoning outcomes. As shown in Table~\ref{tab:case_study}, in the Swahili case, the static baseline (Fix-Merge) fails to retain the critical entity ``60'' and hallucinates an incorrect number ``6 miles'' with a low confidence probability of 0.5523. This indicates that the model was trapped in a state of uncertainty due to the symmetric merging. In contrast, ST-Merge identifies that the input requires prioritizing certain models over others. ST-Merge assigns a higher weight value to the multilingual encoder to boost understanding of the question. Consequently, this asymmetric merging enables the model to generate the correct answer (``60'') with a high confidence of 0.9045, confirming that breaking the symmetry of feature fusion is crucial for robust low-resource reasoning. Similar results can also be observed in the examples of Chinese and French.

\subsection{Analysis of Steerable Weights}

\begin{figure}[!t]
\centering
\begin{tabular}{cc}
    \hspace{-0.35cm} \includegraphics[width=0.48\linewidth]{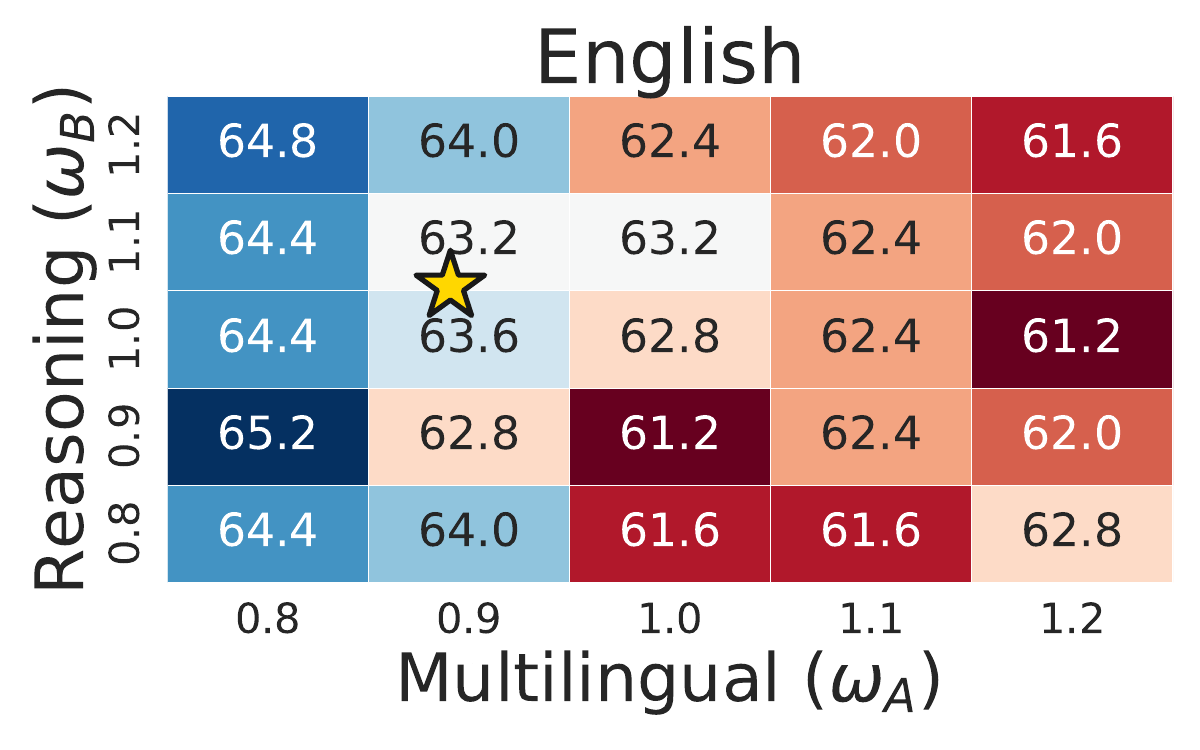} &
    \hspace{-0.35cm} \includegraphics[width=0.48\linewidth]{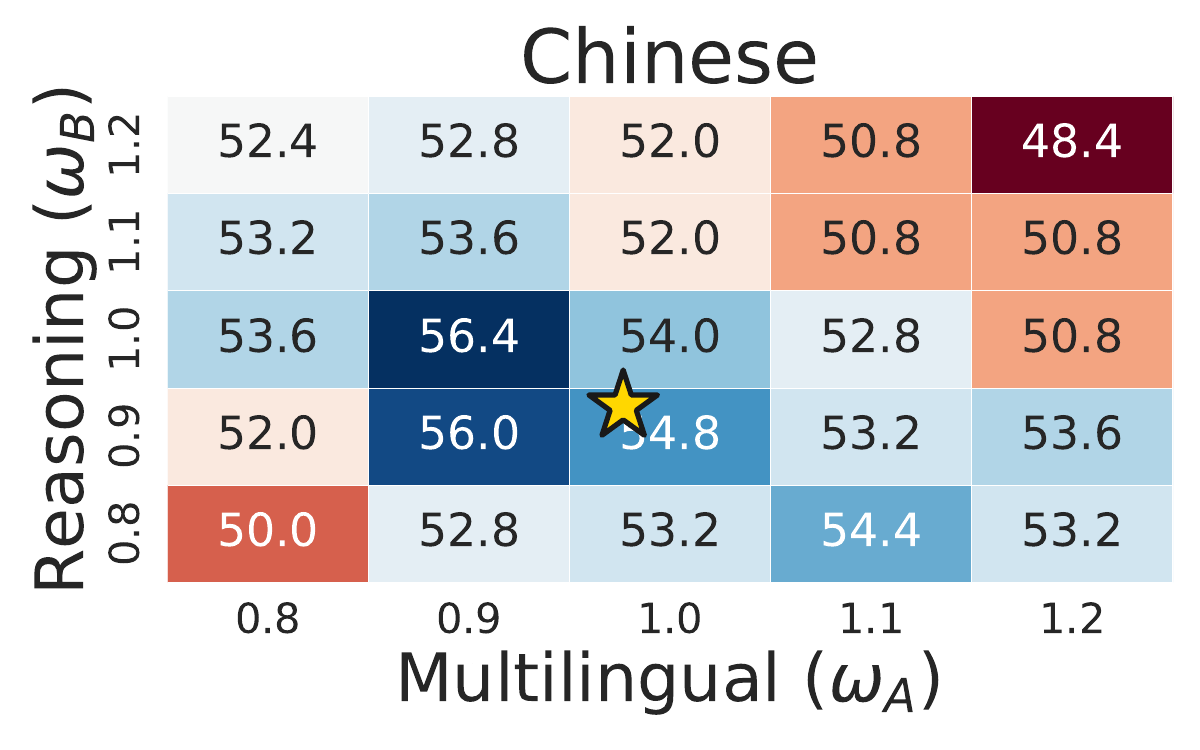}\\
    \hspace{-0.35cm} \includegraphics[width=0.48\linewidth]{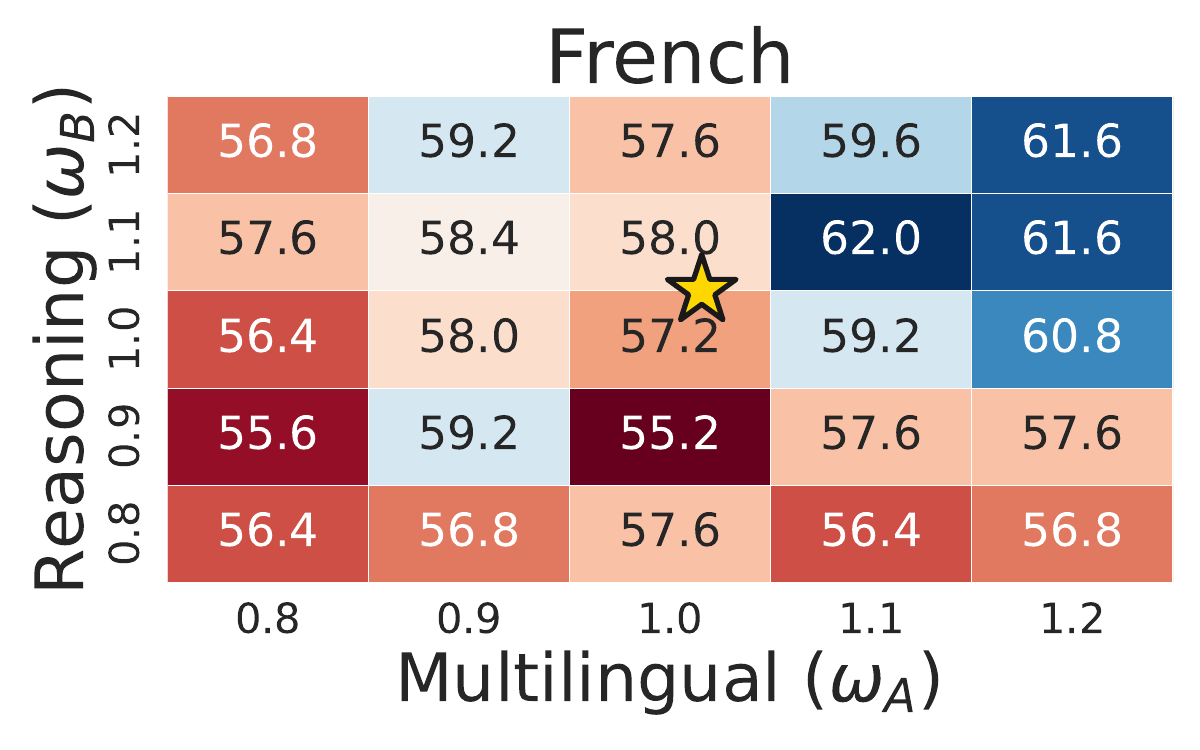} &
    \hspace{-0.35cm} \includegraphics[width=0.48\linewidth]{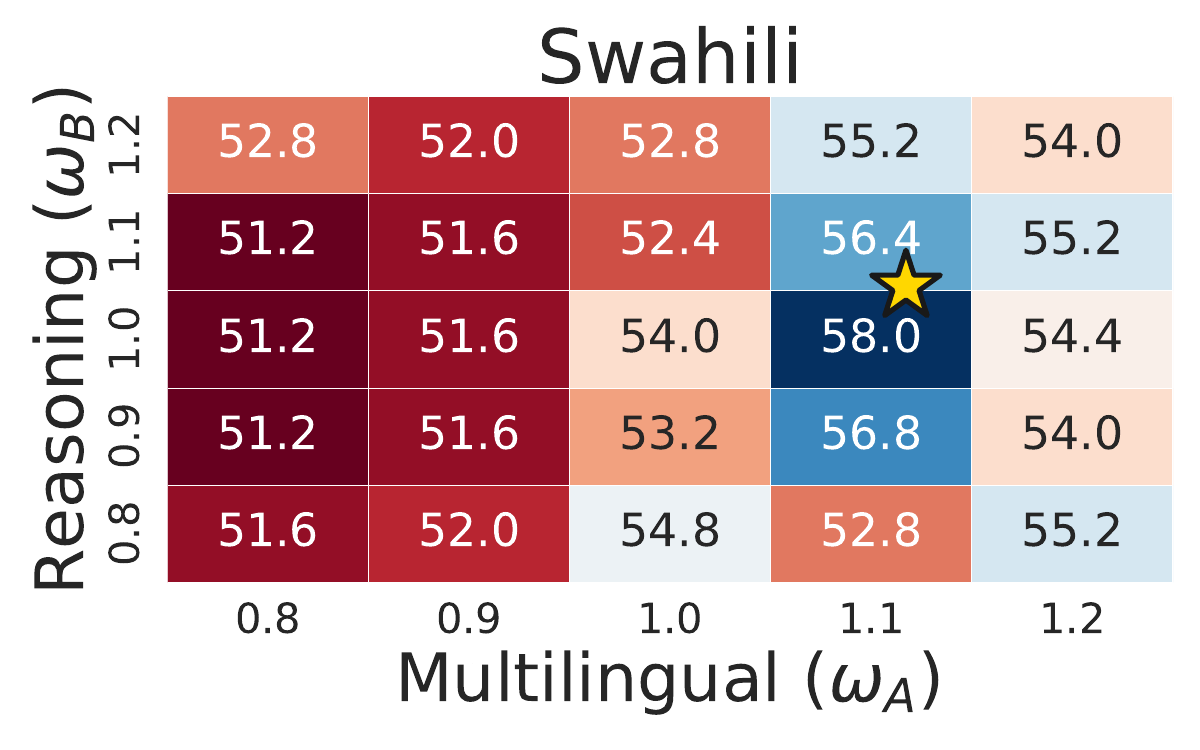}\\
\end{tabular}
\caption{Learned weights analysis of ST-Merge. Darker blue grids indicate higher accuracy. The gold stars represent the learned weights ($\omega_A$, $\omega_B$) by our method for each language.
}
\label{fig:learned_gate}
\end{figure}

Figure~\ref{fig:learned_gate} visualizes the reasoning accuracy on four representative languages from the MGSM dataset (English, Chinese, French, and Swahili) under varying fusion weights $\omega_A$ and $\omega_B$. The heatmaps reveal that the optimal weight configuration is highly sensitive to the specific language. For high-resource languages like English and Chinese, the learned weights (marked by gold stars) and optimal regions favor a balanced or reasoning-dominant configuration, leveraging the strong mathematical reasoning capabilities inherent in the base model. In contrast, for a language underrepresented in math reasoning corpora, e.g. Swahili, the model assigns a higher value to $\omega_A$. This suggests that for low-resource languages, the model prioritizes the multilingual module to align representations before performing reasoning. Crucially, our steerable gating mechanism consistently converges to these optimal regions across all languages, demonstrating its ability to adaptively regulate the trade-off between multilingual alignment and mathematical reasoning without manual tuning.

\begin{figure}[!t]
\centering
\begin{tabular}{cc}
    \hspace{-0.35cm} \includegraphics[width=0.48\linewidth]{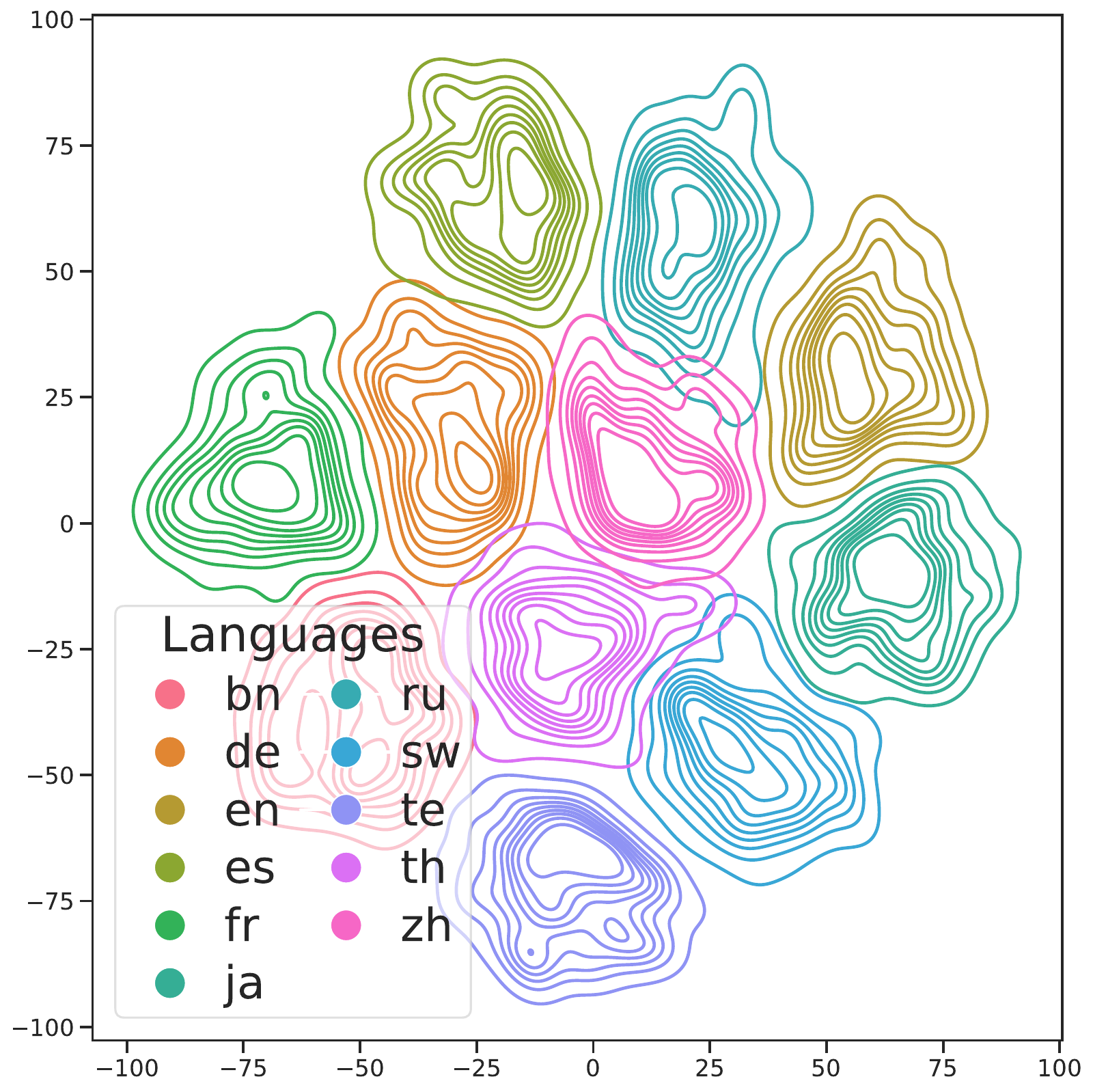} &
    \hspace{-0.35cm} \includegraphics[width=0.48\linewidth]{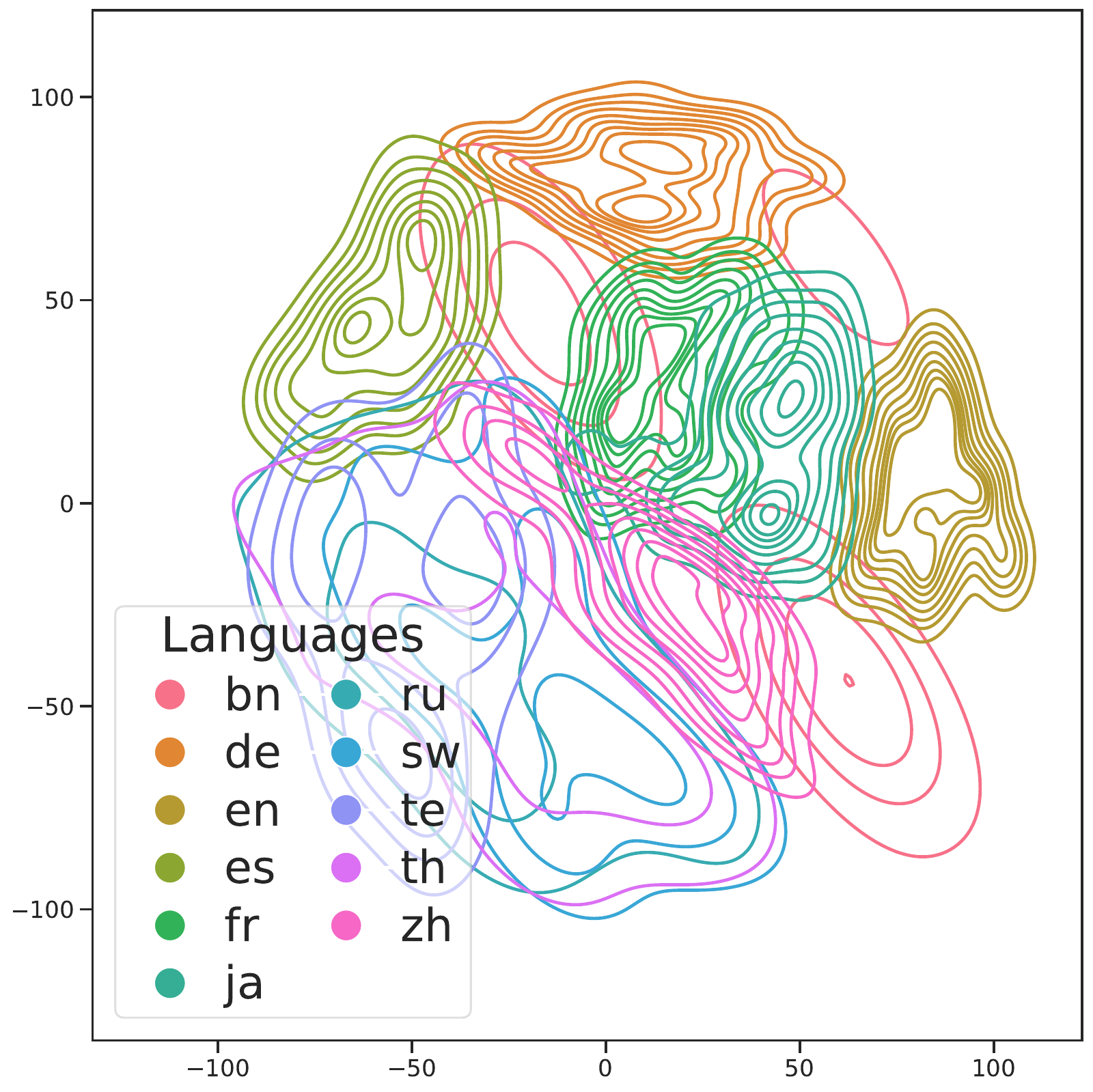} \\
    \hspace{-0.35cm} (a) MetaMath & 
    \hspace{-0.35cm} (b) Our ST-Merge 
\end{tabular}
\caption{t-SNE visualization of multilingual alignment on MGSM.}
\label{fig:tSNE}
\end{figure}

\subsection{Multilingual Representation Visualization}

To examine the alignment of multiple languages, we compare the alignment results of vanilla fine-tuned MetaMath and our ST-Merge in terms of question representation. We select questions spanning eleven different languages from the MGSM datasets  to visualize the embedding space. As
shown in Figure~\ref{fig:tSNE}-(a), different languages are distributed in
distinct clusters in the embedding space, which indicates
that MetaMath can remain highly language-dependent. In contrast, Figure~\ref{fig:tSNE}-(b) shows that the data distributions of different languages are mixed and overlapping, which demonstrates that ST-Merge achieves more effective alignment of representations across languages compared to MetaMath. This alignment contributes to the superior multilingual reasoning performance of ST-Merge.

\section{Conclusion}
\label{sec:conclusion}

This work addresses a fundamental challenge of how to effectively coordinate the multilingual encoder and reasoning LLM for multilingual reasoning tasks. Our analysis reveals that fixed ``one-size-fits-all'' merging strategies potentially introduce a conflict: while improving reasoning performance on low-resource languages with an external multilingual encoder, they often degrade reasoning in high-resource languages where the LLM is already proficient. To address this, we propose a steerable model merging (\textbf{ST-Merge}) framework that optimizes the merged model toward balanced multilingual reasoning via dynamic adjustment of weights. Experiments on four multilingual reasoning benchmarks across 21 languages demonstrate consistent gains across both high-resource and low-resource languages. Beyond performance, we further uncover a correlation between reasoning correctness and gating patterns, providing empirical insight into the mechanisms underlying multilingual reasoning generalization. Additionally, our findings suggest that steerable merging strategies represent a promising direction for enhancing the multilingual capabilities of large language models.

\section*{Limitations}

Our work presents several limitations worth noting. First, to ensure a fair comparison with baseline models, our method primarily conducts experiments using the Llama 2 series models. Future work will involve extending our experiments to additional series models to more comprehensively evaluate the generalizability of our method across diverse backbone architectures. Second, while our method effectively generates weights to improve model collaboration, it lacks fine-grained guidance during the generation process. We hypothesize that a more granular control mechanism during the decoding phase could further enhance performance. In the future, we will explore incorporating token-level or step-aware guidance to address this issue.

\section*{Acknowledgments}
This work is supported by the National Key R\&D Program of China (2024YFF0907003).


\bibliography{custom}

\clearpage
\appendix
\section{Example Appendix}
\label{sec:appendix}

\subsection{Complete Experimental Results}

\begin{table*}[!t]
\small
\setlength{\tabcolsep}{3pt}
\begin{center}
\begin{tabular}{l|cccccccccccccccc|c}
\toprule
\textbf{X-CSQA} & \textbf{Sw} & \textbf{Ur} & \textbf{Hi} & \textbf{Ar} & \textbf{Vi} & \textbf{Ja} & \textbf{Pl} & \textbf{Zh} & \textbf{Nl} & \textbf{Ru} & \textbf{It} & \textbf{De} & \textbf{Pt} & \textbf{Fr} & \textbf{Es} & \textbf{En} & \textbf{Avg.} \\
 \hline

MetaMath~[\citeyear{DBLP:conf/iclr/YuJSYLZKLWL24/metamath}] & 24.2 & 25.1 & 32.9 & 32.3 & 50.9 & 49.1 & 50.6 & 56.5 & 57.5 & 56.0 & 56.0 & 61.2 & 61.7 & 63.5 & 64.0 & 76.3 & 51.3\\

MultiReason~[\citeyear{zhu-etal-2024-question/qalign}] & 27.6 & 29.2 & 32.0 & 28.7 & 38.8 & 38.7 & 45.5 & 43.8 & 45.9 & 46.5 & 50.2 & 49.1 & 51.2 & 52.1 & 54.3 & 67.2 & 43.8\\
QAlign~[\citeyear{zhu-etal-2024-question/qalign}] & 35.1 & 32.6 & 37.8 & 36.3 & 50.5 & 49.2 & 51.3 & 54.8 & 56.3 & 56.3 & 58.3 & 58.8 & 59.8 & 60.3 & 63.1 & 75.7 & 52.3 \\
LangBridge~[\citeyear{yoon-etal-2024-langbridge}] & 31.8 & 30.5 & 30.6 & 30.6 & 33.3 & 33.9 & 39.8 & 39.8 & 38.4 & 35.1 & 39.1 & 37.4 & 36.3 & 38.2 & 38.4 & 44.4 & 36.1 \\
Translate-En~[\citeyear{DBLP:conf/iclr/ShiSF0SVCTRZ0W23/mgsm}] & 36.5 & 41.3 & 48.4 & 44.6 & 51.8 & 47.1 & 53.3 & 51.5 & 55.0 & 54.4 & 56.3 & 57.3 & 54.7 & 57.2 & 55.5 & 71.3 & 52.3 \\
MindMerger~[\citeyear{NEURIPS2024_3bf80b34/mindmerger}] & 45.5 & 46.2 & 48.4 & 51.4 & 60.6 & 53.9 & 63.3 & 62.9 & 63.8 & \textbf{63.7} & \textbf{66.8} & \textbf{67.0} & \textbf{67.1} & \textbf{68.1} & \textbf{69.1} & \textbf{78.1} & 61.0 \\
LayAlign~[\citeyear{ruan-etal-2025-layalign}] & 53.3 & 51.7 & \textbf{53.7} & 55.9 & \textbf{62.0 }& 56.4 & 64.8 & \textbf{64.6} & \textbf{66.2} & 62.0 & 66.2 & 65.2 & 64.3 & 66.5 & 67.3 & 76.7 & 62.3\\
 \rowcolor{gray!20}

 \textbf{$\text{ST-Merge (Ours)}$} & \textbf{53.6} & \textbf{51.9} & {51.6} & 56.5 & {61.7} & \textbf{57.9} & \textbf{65.1} & {64.1} & {64.4} & 63.1 & 66.6 & 65.2 & 66.4 & 67.2 & 67.3 & 77.3& \textbf{62.5}\\
 
\bottomrule
\end{tabular}
\end{center}
\caption{Accuracy (\%) on X-CSQA. Avg. represents the average accuracy across all languages.}
\label{app_tab:main-xcsqa}
\end{table*}

\begin{table*}[!t]
\small
\setlength{\tabcolsep}{3pt}

\begin{center}
\begin{tabular}{l|ccccccccccccccc|c}
\toprule
\textbf{XNLI} & \textbf{Sw} & \textbf{Ur} & \textbf{Hi} & \textbf{Th} & \textbf{Ar} & \textbf{Tr} & \textbf{El} & \textbf{Vi} & \textbf{Zh} & \textbf{Ru} & \textbf{Bg} & \textbf{De} & \textbf{Fr} & \textbf{Es} & \textbf{En} & \textbf{Avg.} \\
 \hline

MetaMath~[\citeyear{DBLP:conf/iclr/YuJSYLZKLWL24/metamath}]  & 45.9 & 49.2 & 55.7 & 55.4 & 60.9 & 61.9 & 63.7 & 73.7 & 74.7 & 77.6 & 76.7 & 80.6 & 82.2 & 82.8 & \textbf{90.0} & 68.7\\

MultiReason~[\citeyear{zhu-etal-2024-question/qalign}]  & 56.3 & 57.5 & 61.7 & 60.1 & 61.7 & 65.6 & 67.0 & 73.7 & 79.1 & 79.7 & 78.7 & 82.3 & 82.9 & 83.9 & 88.8 & 71.9\\
QAlign~[\citeyear{zhu-etal-2024-question/qalign}] & 65.2 & 62.2 & 63.3 & 65.2 & 67.0 & 67.9 & 66.5 & 73.7 & 76.6 & 79.2 & 79.4 & 80.9 & 83.1 & 83.8 & 89.1 & 73.5\\
LangBridge~[\citeyear{yoon-etal-2024-langbridge}] & 71.7 & 66.9 & 71.1 & 72.4 & 75.2 & 74.8 & 79.1 & 78.5 & 77.4 & 77.4 & 79.6 & 78.8 & 79.9 & 80.5 & 83.4 & 76.5 \\
Translate-En~[\citeyear{DBLP:conf/iclr/ShiSF0SVCTRZ0W23/mgsm}] & 65.3 & 61.6 & 68.7 & 69.5 & 68.9 & 74.5 & 79.3 & 76.7 & 74.8 & 76.0 & 80.8 & 80.6 & 80.4 & 81.4 & 87.4 & 75.1 \\
MindMerger~[\citeyear{NEURIPS2024_3bf80b34/mindmerger}] & 66.6 & 69.4 & 74.7 & 71.8 & 76.2 & 75.7 & 78.5 & {80.3} & 80.0 & 80.7 & 82.4 & {83.5} & 83.9 & 84.4 & 88.7 & 78.4 \\
LayAlign~[\citeyear{ruan-etal-2025-layalign}] & 73.0 & 71.0 & 74.7 & 74.1 & \textbf{77.6} & 76.0 & 79.6 & \textbf{80.8} & 80.8 & 81.8 & \textbf{83.4} & \textbf{83.9} & \textbf{84.7} & \textbf{84.8} & 88.9 & 79.7 \\
 \rowcolor{gray!20}

 \textbf{$\text{ST-Merge (Ours)}$} & \textbf{73.7} & \textbf{71.8} & \textbf{75.1} & \textbf{74.2} & \textbf{77.6} & \textbf{77.2} & \textbf{80.0} & 80.1 & \textbf{81.0} & \textbf{82.2} & {83.3} & {83.5} & \textbf{84.7} & \textbf{84.8} & {89.1} & \textbf{79.9} \\
 \bottomrule
\end{tabular}
\end{center}
\caption{Accuracy (\%) on XNLI. Avg. represents the average accuracy across all languages.}
\label{app_tab:main-xnli}
\end{table*}

\begin{table*}[!t]
\centering
\begin{tabular}{lccccc}
\toprule
\textbf{Method} & \textbf{Params (M)} & \textbf{FLOPs (G)} & \textbf{Train (h)} & \textbf{Infer (m)} & \textbf{Avg. Acc} \\ 
\midrule
\textbf{ST-Merge (Ours) }& 10282.5 & 17700.78 & 5.57 & 26.3 & \textbf{60.0} \\ 
\textit{\quad w/o Gate Network}  & 10265.2 & 17700.50 & 4.98 & 26.3 & 57.3 \\
\midrule
\textit{Relative Overhead} & +0.17\% & +0.0016\% & +11.85\% & 0.00\% & +4.71\% \\
\bottomrule
\end{tabular}
\caption{Comparison of computational overhead and performance.}
\label{tab:overhead}
\end{table*}

\begin{figure}[!htbp]
\centering
\begin{tabular}{ccc}
    \hspace{-0.35cm} \includegraphics[width=0.33\linewidth]{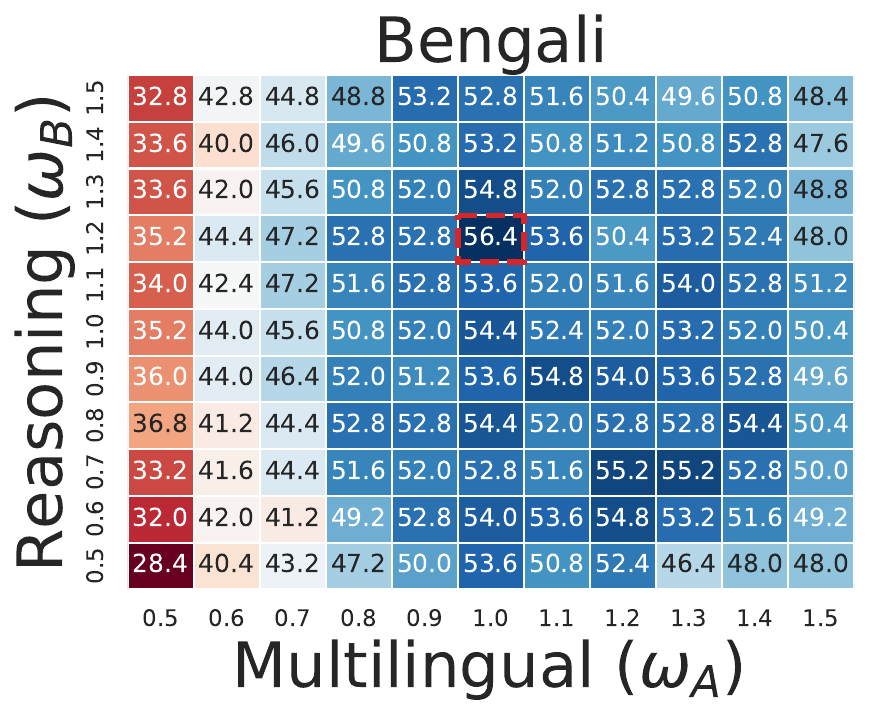} &
    \hspace{-0.35cm} \includegraphics[width=0.33\linewidth]{imgs/region_box_Chinese.pdf} & 
    \hspace{-0.35cm} \includegraphics[width=0.33\linewidth]{imgs/region_box_English.pdf}    \\
    \hspace{-0.35cm} \includegraphics[width=0.33\linewidth]{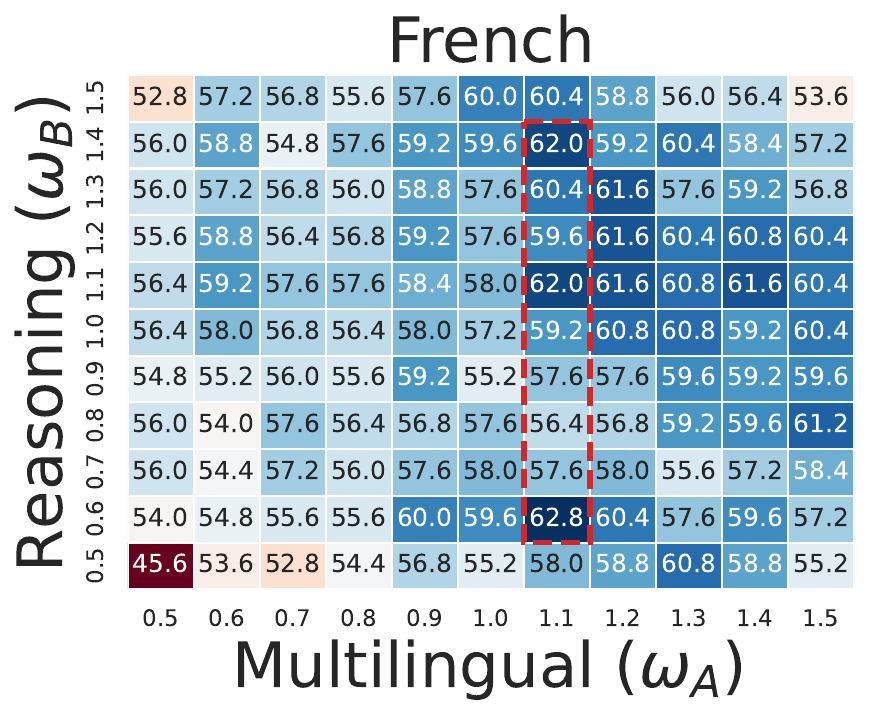} &
    \hspace{-0.35cm} \includegraphics[width=0.33\linewidth]{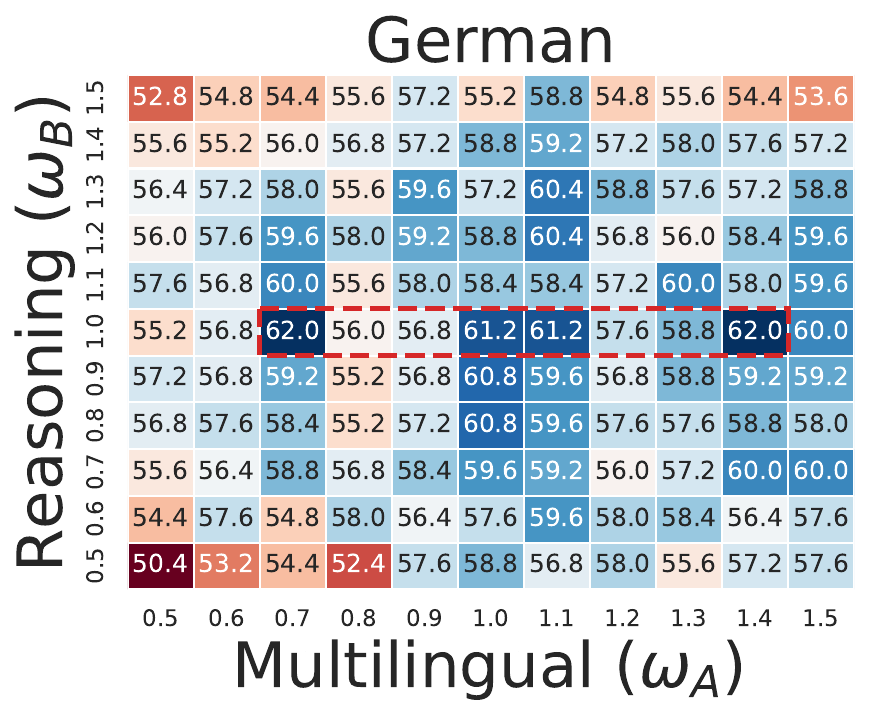} &  
    \hspace{-0.35cm} \includegraphics[width=0.33\linewidth]{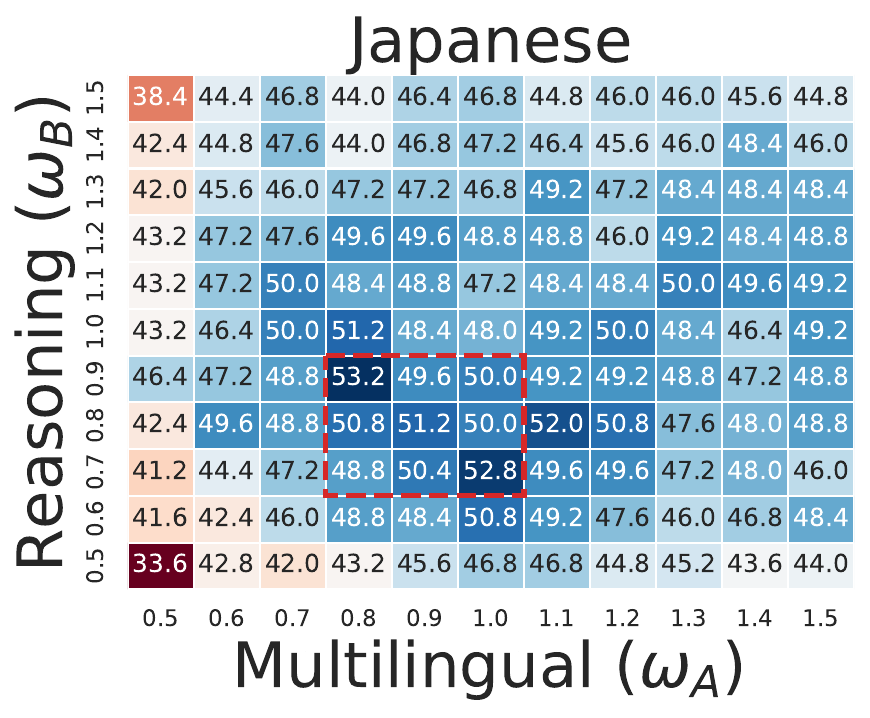} \\
    \hspace{-0.35cm} \includegraphics[width=0.33\linewidth]{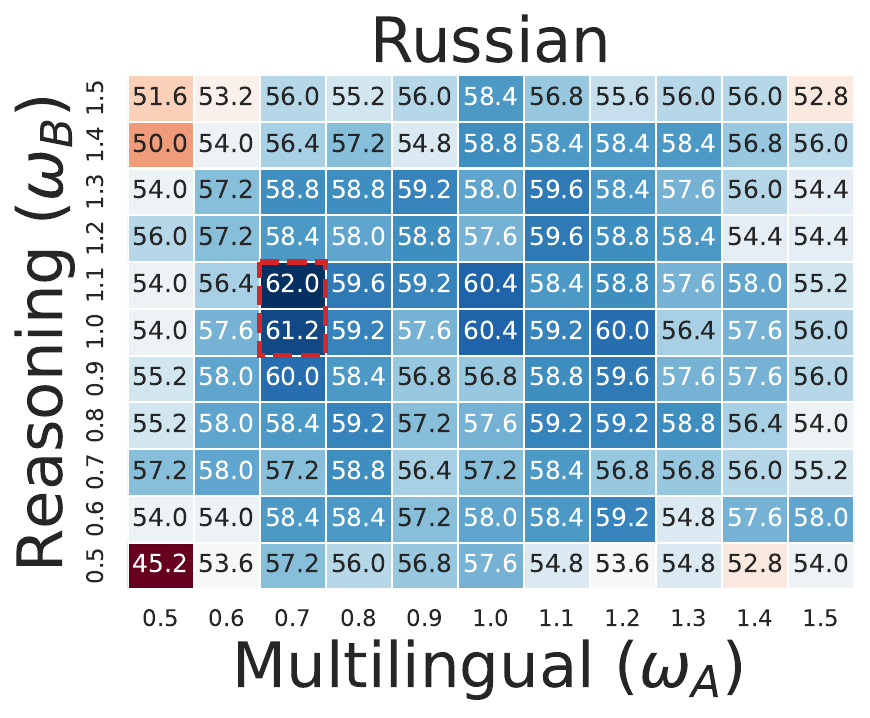} &
    \hspace{-0.35cm} \includegraphics[width=0.33\linewidth]{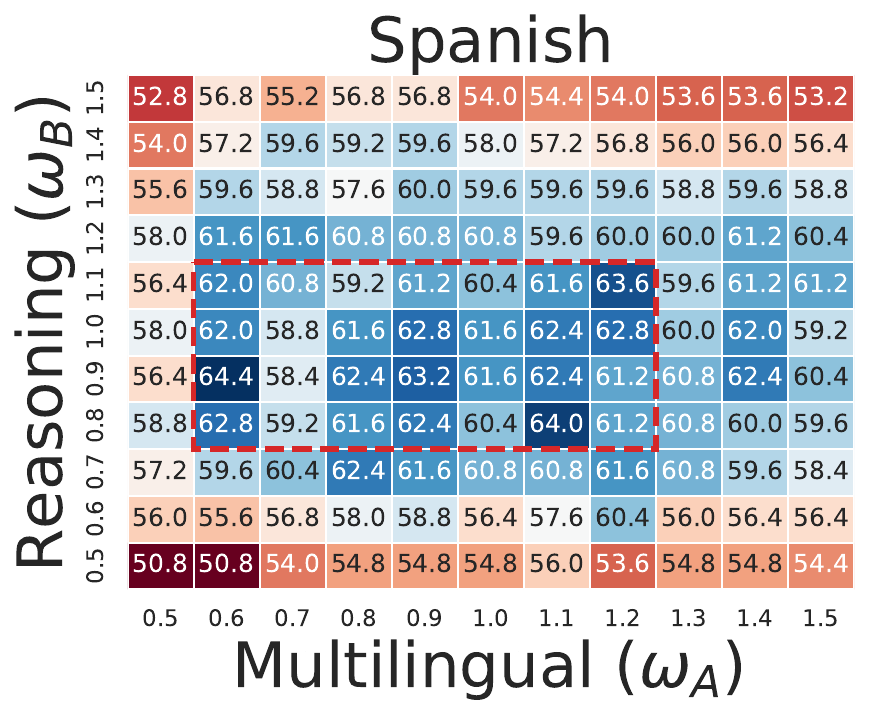} &  
    \hspace{-0.35cm} \includegraphics[width=0.33\linewidth]{imgs/region_box_Swahili.pdf}  \\ 
    \hspace{-0.35cm} \includegraphics[width=0.33\linewidth]{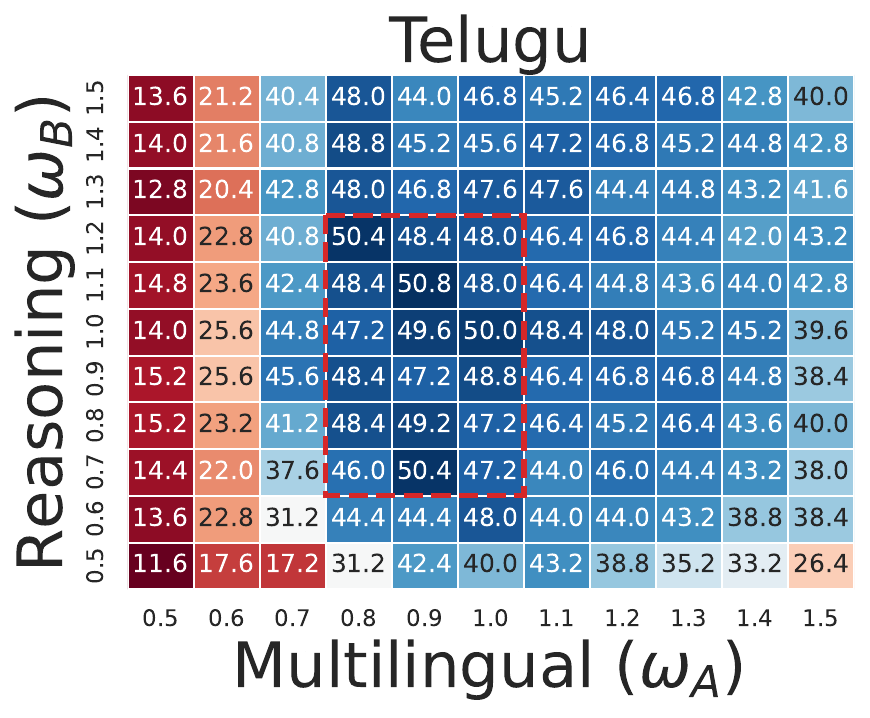} &
    \hspace{-0.35cm} \includegraphics[width=0.33\linewidth]{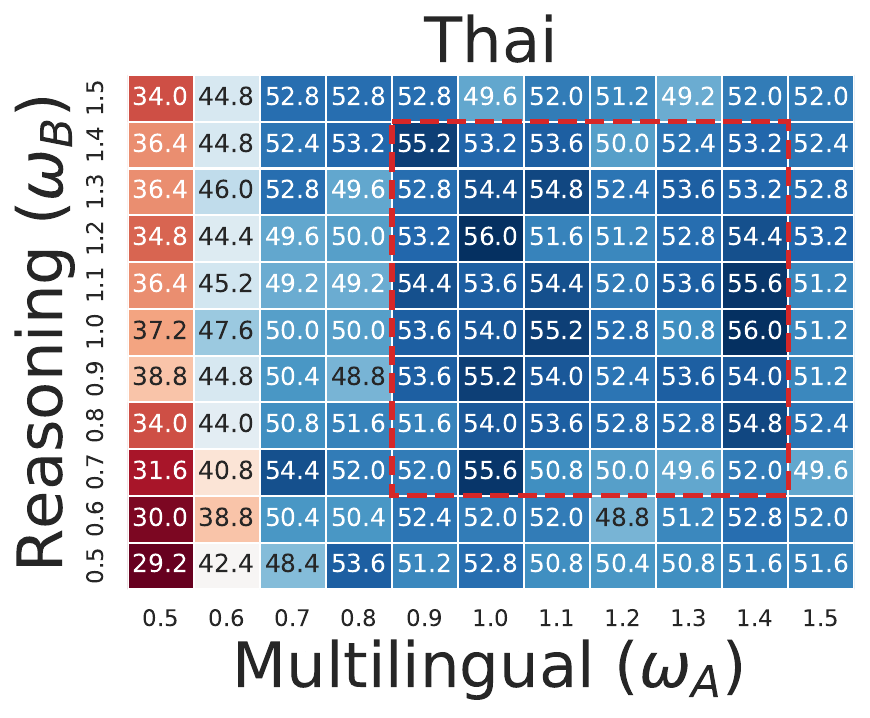}
\end{tabular}
\caption{Accuracy on MGSM with different manual weights combinations for the two source models. Darker blue grids indicate higher reasoning accuracy.}
\label{app_fig:manual_weight}
\end{figure}

\begin{figure}[!htbp]
\centering
\begin{tabular}{ccc}
    \hspace{-0.35cm} \includegraphics[width=0.33\linewidth]{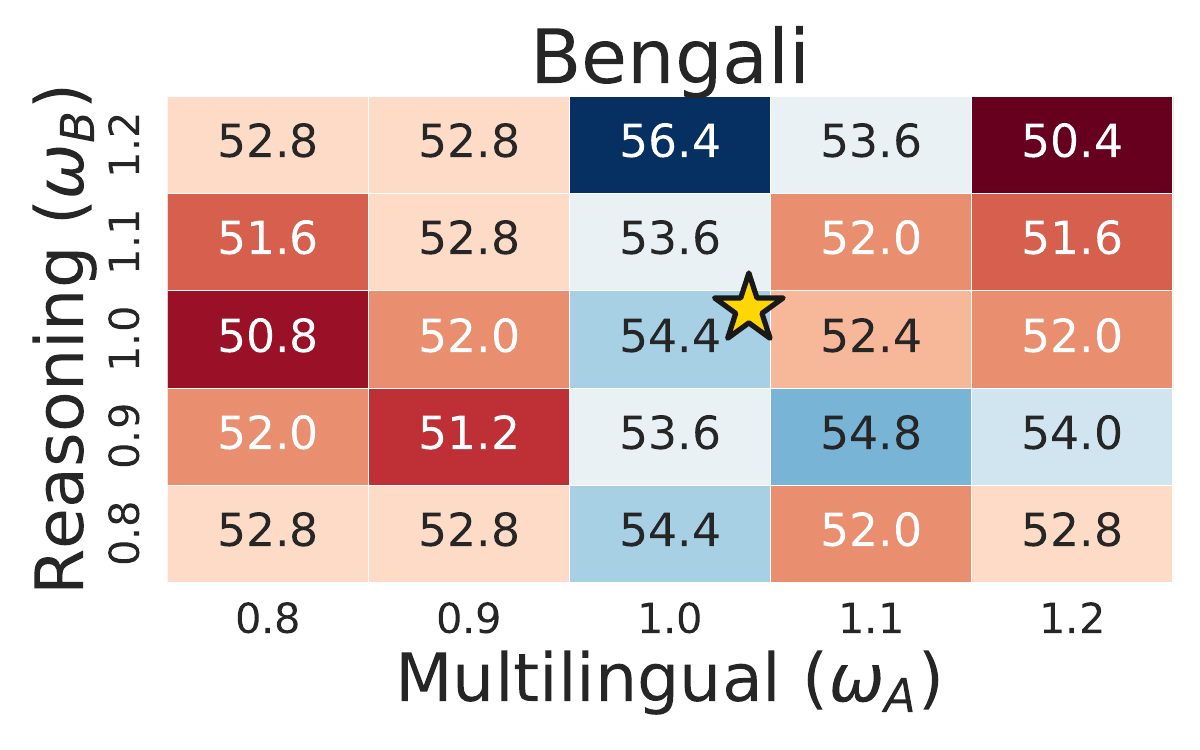} &
    \hspace{-0.35cm} \includegraphics[width=0.33\linewidth]{imgs/heatmap_Chinese.pdf} &
    \hspace{-0.35cm} \includegraphics[width=0.33\linewidth]{imgs/heatmap_English.pdf} \\
    \hspace{-0.35cm} \includegraphics[width=0.33\linewidth]{imgs/heatmap_French.pdf} &
    \hspace{-0.35cm} \includegraphics[width=0.33\linewidth]{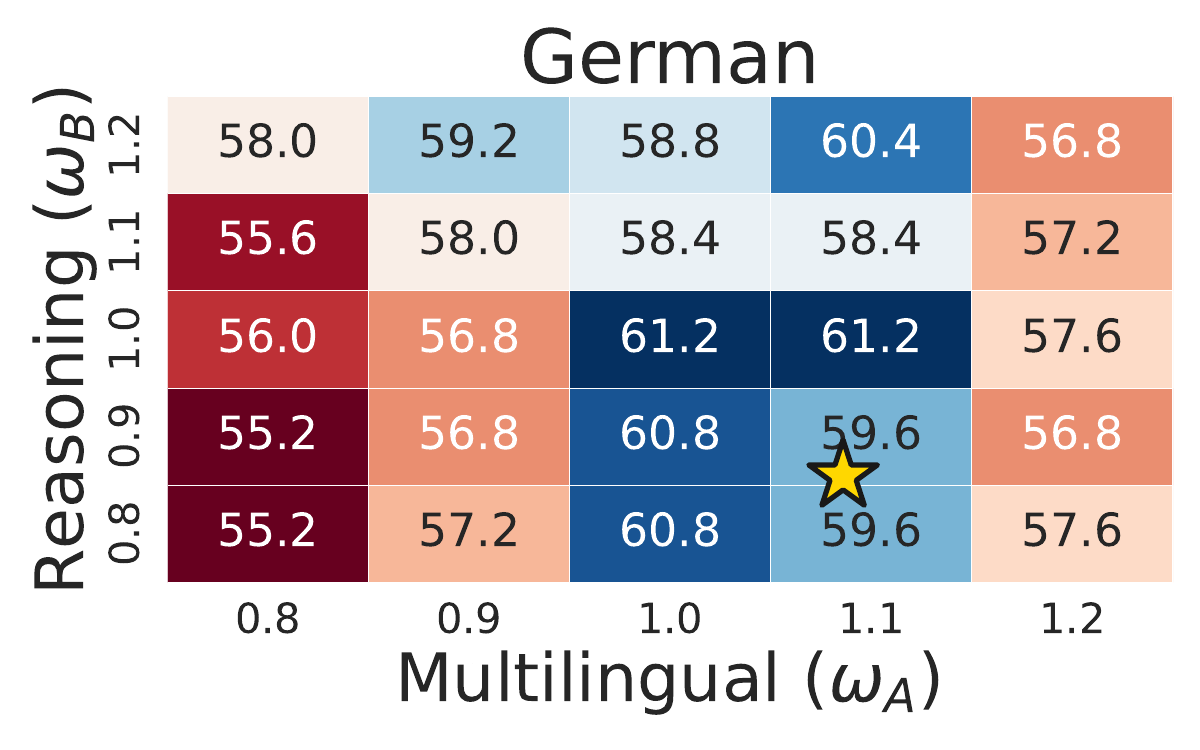} &
    \hspace{-0.35cm} \includegraphics[width=0.33\linewidth]{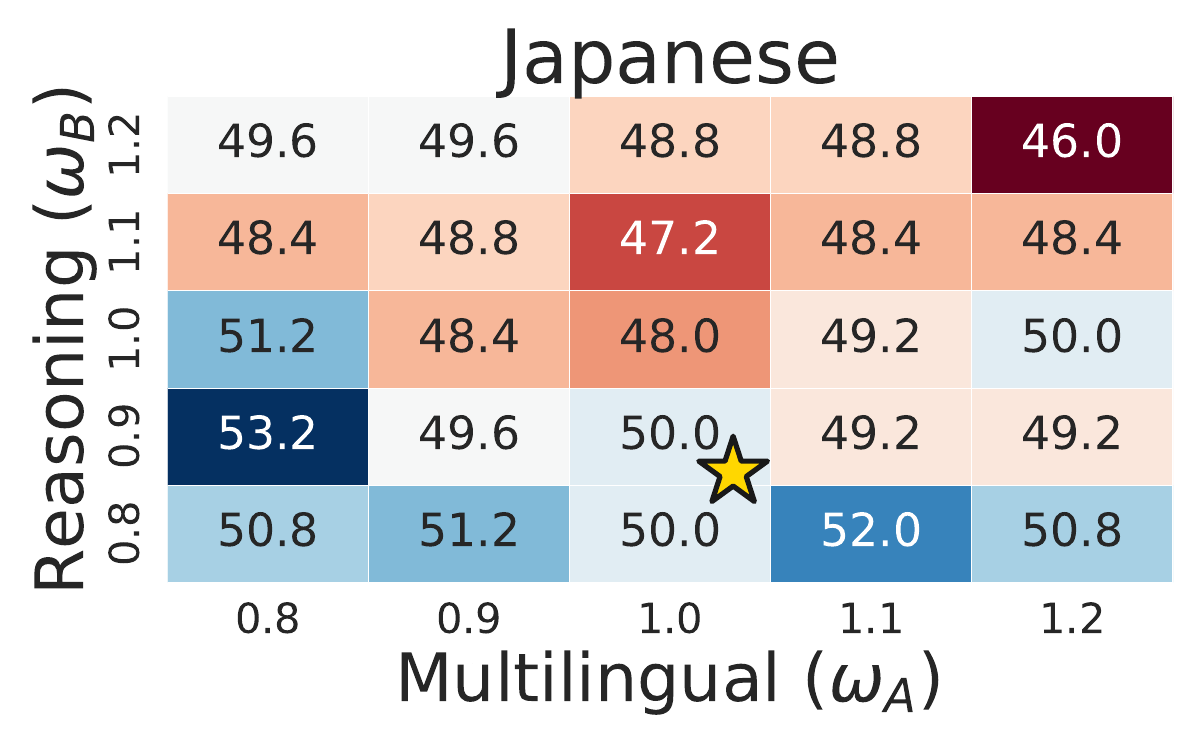}\\
    \hspace{-0.35cm} \includegraphics[width=0.33\linewidth]{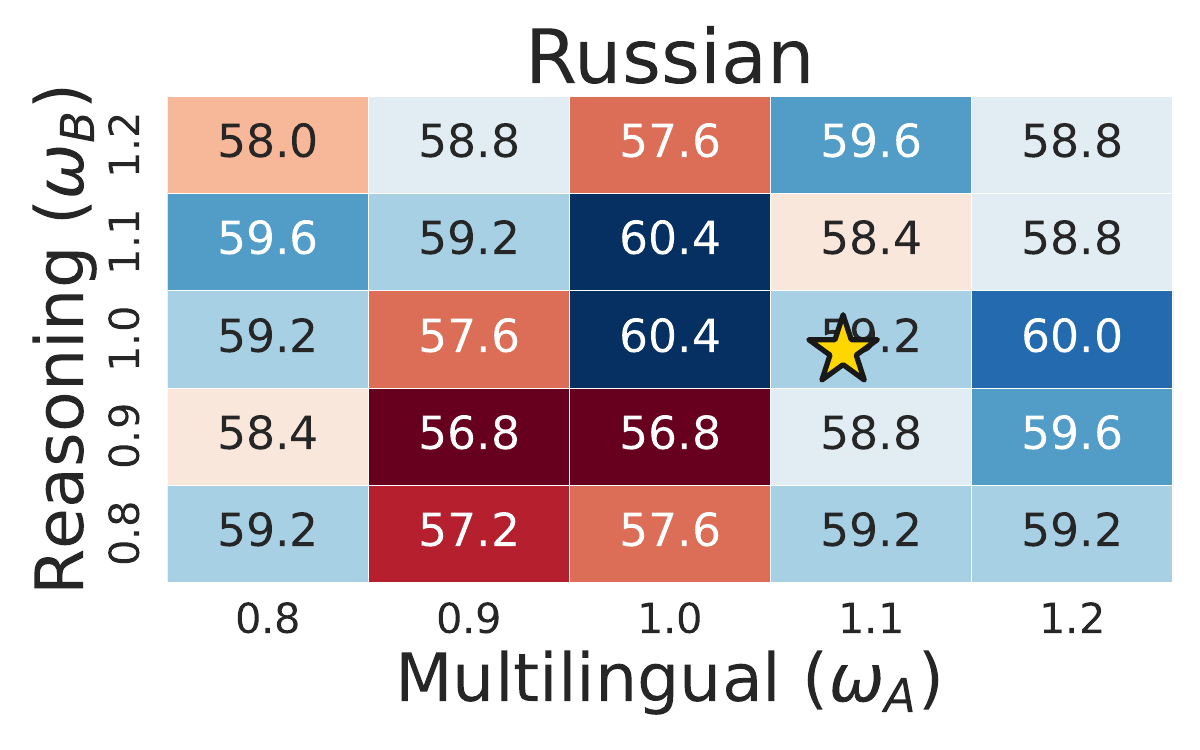} &
    \hspace{-0.35cm} \includegraphics[width=0.33\linewidth]{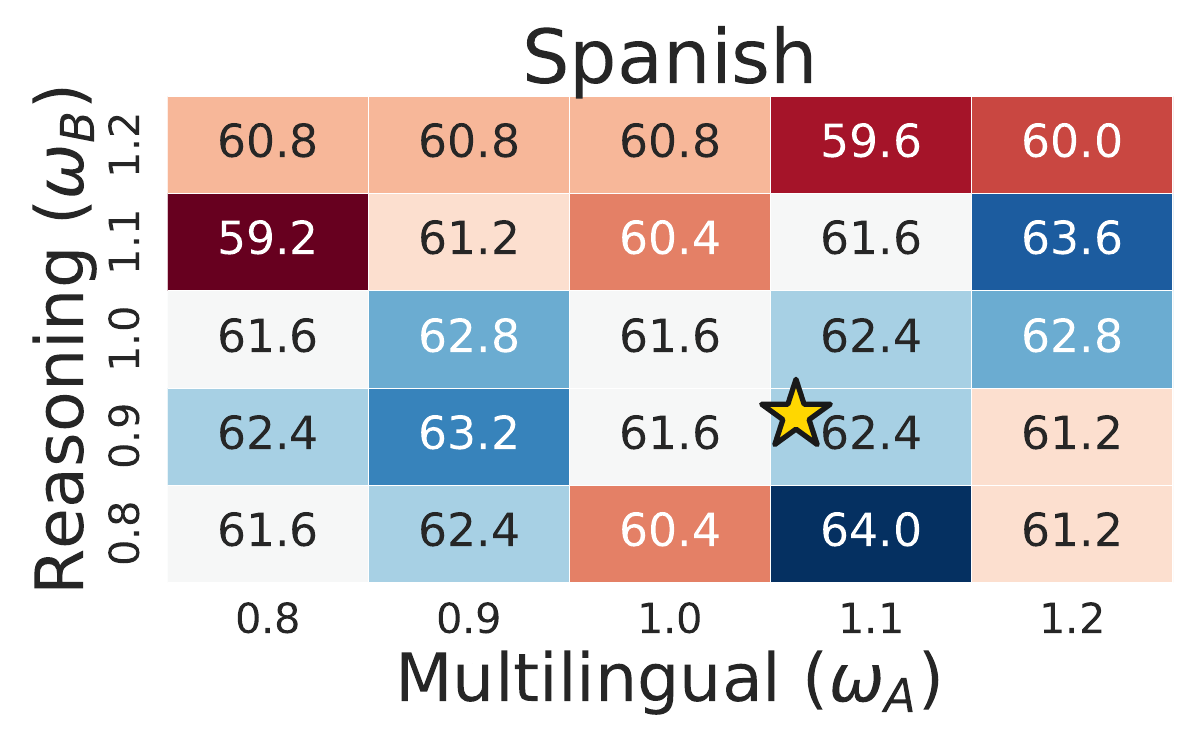} &
    \hspace{-0.35cm} \includegraphics[width=0.33\linewidth]{imgs/heatmap_Swahili.pdf}  \\
    \hspace{-0.35cm} \includegraphics[width=0.33\linewidth]{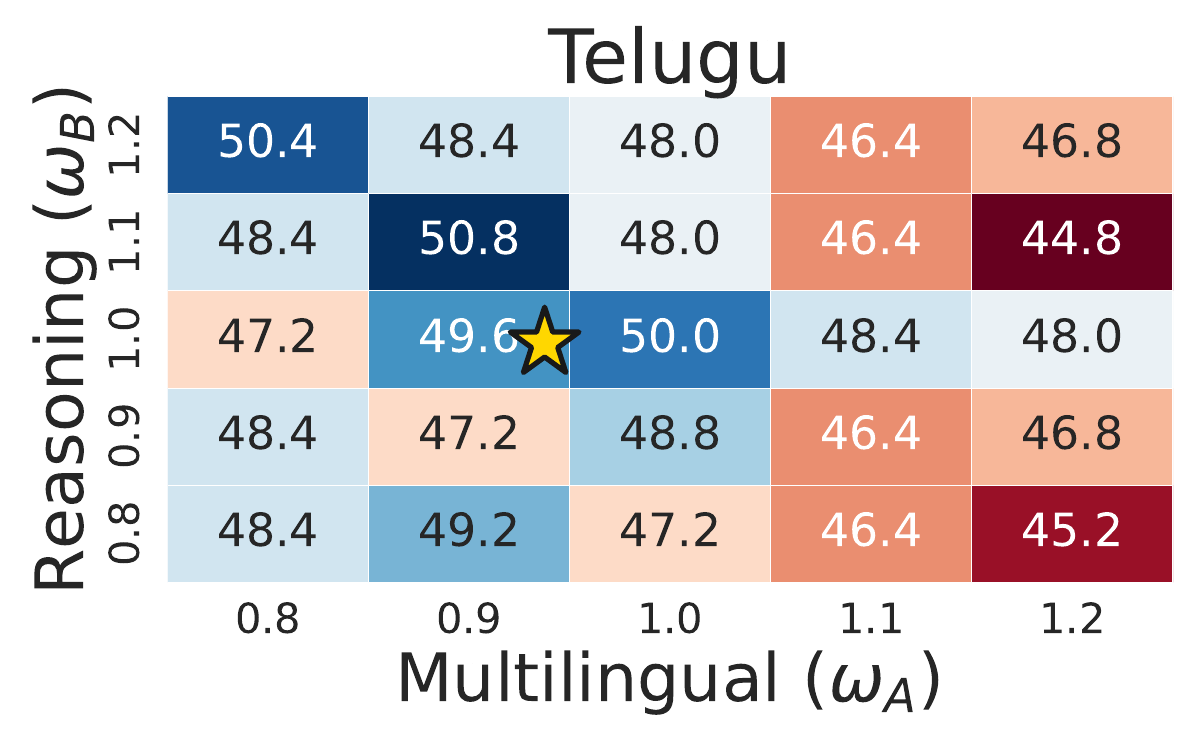} &
    \hspace{-0.35cm} \includegraphics[width=0.33\linewidth]{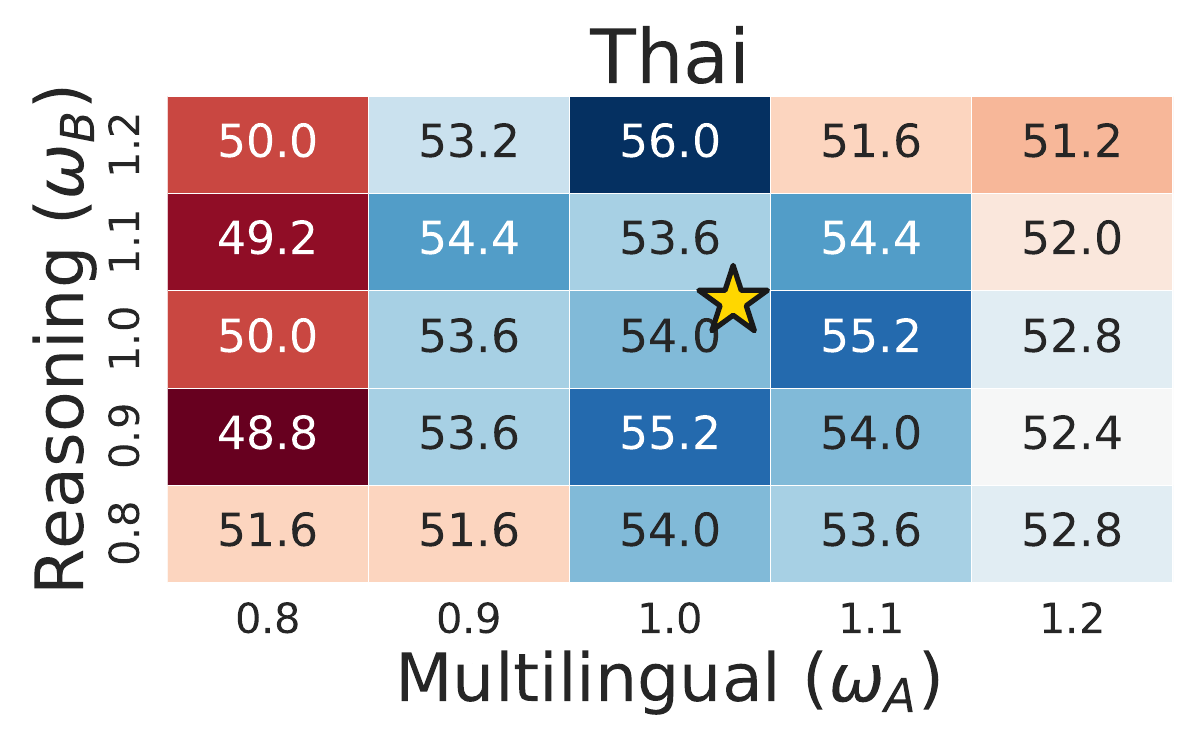}
\end{tabular}
\caption{Learned weights analysis of ST-Merge. Darker blue grids indicate higher accuracy. The gold stars represent the learned weights ($\omega_A$, $\omega_B$) by our method for each language.
}
\label{app_fig:learned_gate}
\end{figure}

To facilitate reference, the languages utilized in this work are abbreviated as follows: Bengali (Bn), Thai (Th), Swahili (Sw), Japanese (Ja), Chinese (Zh), German (De), French (Fr), Russian (Ru), Spanish (Es), English (En), Urdu (Ur), Hindi (Hi), Arabic (Ar), Vietnamese (Vi), Polish (Pl), Flemish (Nl), Italian (It), Portuguese (Pt), Turkish (Tr), Greek (El), and Bulgarian (Bg). Due to page limitations, the complete breakdown of results is included here. We report the extensive experimental data on X-CSQA in Table~\ref{app_tab:main-xcsqa} and on XNLI in Table~\ref{app_tab:main-xnli}. Additionally, comparative results on MGSM are visualized in Figure~\ref{app_fig:manual_weight} (manual weights combinations) and Figure~\ref{app_fig:learned_gate} (learned weights combinations).

\subsection{Computational Overhead Analysis}
To verify whether the performance gains of ST-Merge stem from our proposed steerable merging design rather than a simple increase in parameter capacity, we evaluate its computational overhead against the \textit{ w/o Gate Network} variant. As summarized in Table \ref{tab:overhead}, ST-Merge introduces only a marginal parameter increase of 17.3M (+0.17\%) and a negligible overhead in FLOPs (+0.0016\%). Although the training time increases by 11.85\%, the inference latency remains identical to the baseline. Considering the significant absolute improvement of 2.7 points (+4.71\%) in average accuracy, these results demonstrate that the effectiveness of ST-Merge is driven by its architectural contribution rather than parameter expansion, ensuring its practicality for resource-constrained applications.

\end{document}